\pdfoutput=1

\documentclass[11pt]{article}
\PassOptionsToPackage{table}{xcolor}

\usepackage[final]{acl}

\usepackage{times}
\usepackage{latexsym}
\usepackage{hyperref}
\usepackage{url}
\usepackage{booktabs}
\usepackage{multirow}
\usepackage{bm}
\usepackage[table]{xcolor}
\usepackage{graphicx}
\usepackage{subcaption}
\usepackage{algorithm}
\usepackage{algpseudocode}
\usepackage{tabularx}
\usepackage{listings}
\usepackage{amsmath}
\usepackage{amssymb}
\usepackage{colortbl}

\usepackage[T1]{fontenc}

\usepackage[utf8]{inputenc}

\usepackage{microtype}

\usepackage{inconsolata}

\title{AdaLomo: Low-memory Optimization with Adaptive Learning Rate}

\author{
Kai Lv\textsuperscript{1,2}\thanks{Work done during internship at Shanghai AI Laboratory.}, 
Hang Yan\textsuperscript{2}, 
Qipeng Guo\textsuperscript{2}\thanks{Corresponding author.},
Haijun Lv\textsuperscript{2}, 
Xipeng Qiu\textsuperscript{1}\\
\textsuperscript{1}School of Computer Science, Fudan University\\
\textsuperscript{2}Shanghai AI Laboratory\\
\texttt{klv21@m.fudan.edu.cn}, \texttt{\{yanhang,guoqipeng,lvhaijun\}@pjlab.org.cn}\\
\texttt{xpqiu@fudan.edu.cn}
}

\begin{document}
\maketitle
\begin{abstract}
Large language models have achieved remarkable success, but their extensive parameter size necessitates substantial memory for training, thereby setting a high threshold. While the recently proposed low-memory optimization (LOMO) reduces memory footprint, its optimization technique, akin to stochastic gradient descent, is sensitive to hyper-parameters and exhibits suboptimal convergence, failing to match the performance of the prevailing optimizer for large language models, AdamW. Through analysis of the Adam optimizer, we found that, compared to momentum, the adaptive learning rate is more critical for bridging the gap. Building on this insight, we introduce the low-memory optimization with adaptive learning rate (AdaLomo), which offers an adaptive learning rate for each parameter and exhibits superior convergence performance compared to LOMO theoretically. To maintain memory efficiency, we employ non-negative matrix factorization for the second-order moment estimation. Additionally, we suggest the use of a grouped update normalization to stabilize convergence. Our experiments on instruction-tuning, further pre-training and from-scratch pre-training demonstrate that AdaLomo achieves results on par with AdamW, while significantly reducing memory requirements, thereby lowering the hardware barrier to training large language models. The code is accessible at \url{https://github.com/OpenLMLab/LOMO}.
\end{abstract}

\section{Introduction}
Large language models~\citep{bloom,opt,llama,llama2} have garnered increasing attention due to their exceptional capabilities across a diverse range of tasks. 
Either supervised fine-tuning or further pre-training can lead to enhanced performance.
As the number of parameters grows, the substantial GPU memory required for training sets a high hardware threshold. Recently, \citet{lomo} has proposed low-memory optimization (LOMO) to train large language models in a memory-saving approach by simultaneously backpropagating gradients and updating parameters during the backward pass, enabling the fine-tuning of all parameters of a 7B model on a consumer-grade RTX 3090.

While LOMO's performance on the SuperGLUE~\citep{superglue} benchmark is comparable to popular parameter-efficient fine-tuning methods~\citep{peft_overview,lora}, it falls short on a broader range of tasks against adaptive optimization methods like Adam~\citep{adam}, exhibiting a convergence gap. We attribute this to its reliance on the naive stochastic gradient descent optimization approach. 
We analyze the differences in optimization methods between Adam and LOMO. Compared to LOMO, Adam incorporates both the first and second moment estimation in its optimizer state, which are the moving averages of the gradient and the squared gradient, respectively. Based on our theoretical and empirical analysis, we identify that the second moment estimation is the pivotal factor influencing the convergence of training large language models between LOMO and Adam.
\begin{table*}[t]
\small
\centering
\begin{tabular}{lccccc}
\toprule
\multirow{2}{*}{\textbf{Method}} & \textbf{Trainable Params} & \multicolumn{4}{c}{\textbf{Memory (GB)}}                          \\ \cline{3-6} 
                                 & \textbf{(Billion)}        & Param  & Gradient     & Optimizer State & Total                \\ \midrule
LoRA                             & $N$                & $2M$ & $O(N)$ & $O(N)$    & $\bm{\sim2M}$ \\
AdamW                            & $\bm{M}$          & $2M$ & $2M$       & $12M$         & $16M$             \\ \rowcolor[gray]{0.9}
AdaLomo                          & $\bm{M}$          & $2M$ & $O(N)$ & $O(N)$    & $\bm{\sim2M}$ \\ \bottomrule
\end{tabular}
\caption{Trainable parameter number and memory usage under mixed-precision training. \(N \ll M\) and \(O(M+N)=O(M)\), where \(M\) is the number of model parameters. AdaLomo's memory consumption is comparable to LoRA, and its trainable parameter number is equivalent to AdamW.}
\label{tab:compare_param}
\end{table*}

The second-order moment estimation in Adam serves to offer an adaptive learning rate for each parameter. Expanding on this concept, we introduce the low-memory optimization with adaptive learning rate (AdaLomo), which similarly provides an adaptive learning rate for each parameter, thus exhibiting superior convergence performance compared to LOMO theoretically.
To retain memory efficiency, inspired by Adafactor~\citep{adafactor}, we employ non-negative matrix factorization~\citep{matrix_factor} for the second-order moment estimation in the optimizer state. 
We advocate for the use of a grouped update normalization to stabilize convergence instead of global update normalization, which nearly doubles the training speed of AdaLomo while maintaining its performance.
Moreover, under identical conditions, AdaLomo's memory utilization accounts for only approximately 40\% of that consumed by Adafactor.
The number of trainable parameters and the GPU memory consumption for model state under mixed-precision training among AdaLomo, the popular LoRA~\citep{lora} method, and the AdamW optimizer~\citep{adamw} are compared in Table~\ref{tab:compare_param}.

Our contributions are as follows:

\begin{enumerate}
    \item 
    We examined the distinctions between the LOMO and Adam optimization techniques. Analysis in Section~\ref{subsec:anaysis_moments} revealed that the primary difference in performance between LOMO and Adam, especially when training large language models, stems from Adam's incorporation of second-moment estimation.
    \item We introduce AdaLomo, which provides an adaptive learning rate for each parameter while maintaining memory efficiency, democratizing the training of large language models.
    In AdaLomo, we also employ grouped update normalization to stabilize the training process.
    \item 
    We evaluate the performance of large language models post instruction-tuning with AdaLomo across five benchmarks spanning diverse tasks. The results are comparable to both AdamW and LoRA. Furthermore, when AdaLomo is used for pre-training from scratch and further pre-training on Chinese and Python code, its performance is on par with that of AdamW.
    \item We profile the memory consumption and throughput of AdaLomo. Its reduced memory usage and reasonable computational overhead make it a viable option for training large language models.
\end{enumerate}

\section{Preliminaries}
In the subsequent sections of this paper, we use \( \boldsymbol{\theta_t} \) to denote the parameters of the model at the \( t^{th} \) step of the training process. \( \theta_{t,i} \) represents the parameter at the \( i^{th} \) gradient computation during the backpropagation process of \( \boldsymbol{\theta_t} \). We use \( \boldsymbol{g_t} \) to represent the gradient of \( \boldsymbol{\theta_t} \), and \( g_{t,i} \) to denote the gradient of \( \theta_{t,i} \). The first and second moment estimation at the \( t^{th} \) training step, which are the moving averages of the gradient and the square of the gradient respectively, are represented by \( \boldsymbol{m_t} \) and \( \boldsymbol{v_t} \). The symbol \( \alpha \) represents the learning rate.

\subsection{Fused Backward} \label{sec:grad_norm}

In the training process, the memory is primarily consumed by the optimizer states, parameters, and gradients. The fused backward proposed in LOMO refers to the process that simultaneously calculates gradients and updates parameters during backpropagation. This can effectively reduce the memory consumption of gradients.

For a given parameter \(\theta_{t,i}\), its gradient \(g_{t,i}\) resides in the GPU memory until the gradient \(g_{t,i+1}\) corresponding to the subsequent parameter \(\theta_{t,i+1}\) is computed. Subsequently, LOMO utilizes a standard gradient descent approach for parameter updates, as depicted by the following equation:
\begin{gather}
\label{eq:lomo}
\theta_{t,i} = \theta_{t-1,i} - \alpha \times g_{t,i} .
\end{gather}
For transformer-based language models, \(g_{t,i}\) is unnecessary in subsequent backpropagation steps and can be eliminated from memory. Consequently, at any given moment, the memory retains the gradients of only two consecutive parameters. The memory usage for gradients remains constant regardless of the language model's scale, yielding an \(O(1)\) memory footprint. 
In the case of large language models, such as LLaMA-65B~\citep{llama} with its 82 layers and 723 weight matrices, the memory consumption for gradients becomes negligible compared to that for parameters or optimizer states.

\paragraph{Gradient Normalization} 
\begin{figure}[ht]
    \centering
        \includegraphics[width=0.87\linewidth]{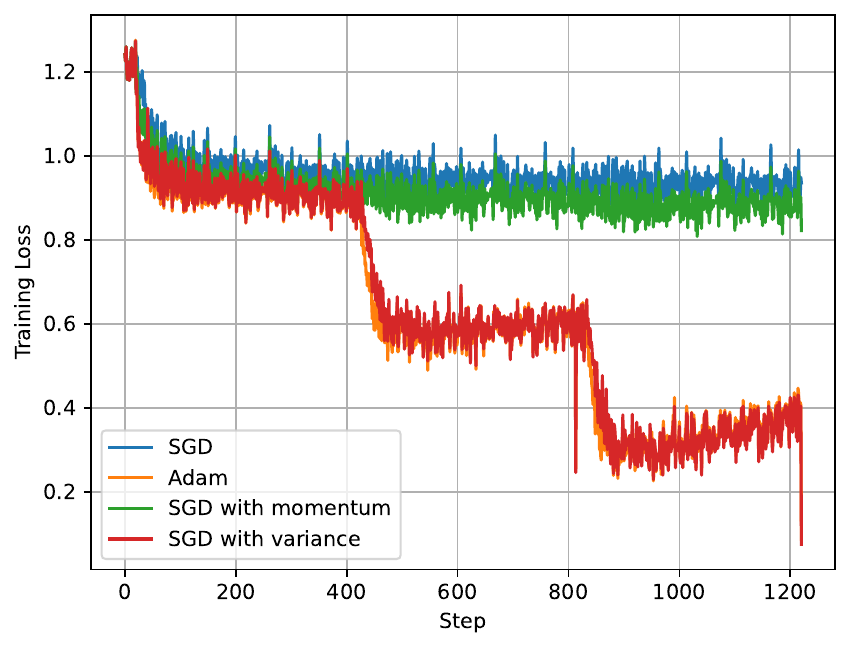}
        \caption{Empirical analysis on different optimization methods. Both Adam and SGD with variance exhibit a stepwise decline in loss as the training epochs increase, ultimately achieving a lower loss than both SGD and SGD with momentum.}
        \label{fig:optim_sft}
\end{figure}
Gradient normalization is instrumental in mitigating gradient vanishing and explosion. Nevertheless, its integration into LOMO presents challenges. Specifically, gradient normalization necessitates the computation of a scaling factor derived from the gradients of all parameters. This factor subsequently informs parameter updates. In the context of LOMO, however, the gradients for all parameters have not yet been computed. To incorporate gradient normalization within LOMO, two backward passes are essential: the first backward pass to get the the overall gradient scaling factor and the second updating the parameters using the derived scaling factor.

This process almost doubles the training time for LOMO. In AdaLomo, we employ grouped update normalization, which requires only a single backward pass to complete.

\subsection{Analysis on the Two Moments in Adam}
\label{subsec:anaysis_moments}

LOMO exhibits efficient memory usage, essentially pushing the optimization of large language models with gradients to the extreme. However, the naive gradient descent method shown in Equation~\ref{eq:lomo} faces challenges such as the propensity to get trapped in saddle points and sensitivity to the learning rate~\citep{saddle_point,lr}. Building upon SGD, a series of advanced optimization methods have been proposed that have been proven both theoretically and practically to address these challenges~\citep{optim_overview}. These methods typically introduce additional optimizer states, such as momentum~\citep{momentum}, Nesterov accelerated gradient~\citep{nesterov_mom}, and moving averages of squared past gradients~\citep{adagrad,adadelta,adam}, leading to extra memory consumption. Among these, the Adam series of optimizers are most widely used in training large language models, simultaneously incorporating first-moment (\( \boldsymbol{m_t} \)) and second-moment (\( \boldsymbol{v_t} \)) estimation for parameter updates, as demonstrated in the following equation, 

\begin{gather}
\left\{
\begin{aligned}
\bm{m}_t & = \beta_1 \bm{m}_{t-1} + (1 - \beta_1) \bm{g}_t ,\\
\bm{v}_t & = \beta_2 \bm{v}_{t-1} + (1 - \beta_2) \bm{g}_t^2 ,\\
\hat{\bm{m}}_t & = \frac{\bm{m}_t}{1 - \beta_1^t} ,\\
\hat{\bm{v}}_t & = \frac{\bm{v}_t}{1 - \beta_2^t} ,\\
\bm{\theta}_t & = \bm{\theta}_{t-1} - \alpha \frac{\hat{\bm{m}}_t}{\sqrt{\hat{\bm{v}}_t} + \epsilon} ,
\end{aligned}
\right.
\end{gather}
where \(\epsilon\) is a small quantity introduced to prevent division by zero in calculations. The hyper-parameters \( \beta_1, \beta_2 \in [0, 1)\) dictate the exponential decay rates of the respective moving averages.

\paragraph{Theoretical Analysis}
\citet{Lipschitz} found that the SGD optimizer is highly sensitive to the network's Lipschitz constant. A significant variance in the Lipschitz constant across different layers results in substantial gradient disparities, leading to inconsistent step sizes in parameter updates with SGD. In contrast, the Adam optimizer employs an adaptive learning rate approach, normalizing update values and demonstrating robustness to variations in the Lipschitz constant. \citet{lipschitz_sa} demonstrated that self-attention structures lack a bounded Lipschitz constant, suggesting that the gradient disparities across different layers in transformer architectures could be significant. Therefore, incorporating an adaptive learning rate into LOMO could enhance optimization for the widely used Transformer architecture~\cite{transformer}.

\paragraph{Empirical Analysis}
We empirically investigated the differences in convergence behaviors between Adam and SGD under the fine-tuning of large language models. To ablatively analyze the roles of the first and second moments of the gradients in Adam, we conducted experiments retaining only the first-order moment estimate or the second-order moment estimation in Adam, respectively.
The update rule retaining only the first-order moment estimation (or momentum) is:
\begin{gather}
\left\{
\begin{aligned}
\bm{m}_t & = \beta_1 \bm{m}_{t-1} + (1-\beta_1) \bm{g}_t ,\\
\hat{\bm{m}}_t & = \frac{\bm{m}_t}{1 - \beta_1^t} ,\\
\bm{\theta}_t & = \bm{\theta}_{t-1} - \alpha \times \hat{\bm{m}}_t .
\end{aligned}
\right.
\end{gather}
Meanwhile, the update rule retaining only the second-order moment estimation (or variance) is:
\begin{gather}
\left\{
\begin{aligned}
\bm{v}_t & = \beta_2 \bm{v}_{t-1} + (1 - \beta_2) \bm{g}_t^2 ,\\
\hat{\bm{v}}_t & = \frac{\bm{v}_t}{1 - \beta_2^t} ,\\
    \bm{\theta}_t & = \bm{\theta}_{t-1} - \alpha \frac{\bm{g}_t}{\sqrt{\hat{\bm{v}}_t} + \epsilon} .
\end{aligned}
\right.
\end{gather}

The results of the convergence analysis are shown in Figure~\ref{fig:optim_sft}. In the instruction-tuning scenario, we trained LLaMA-7B~\citep{llama} with the Alpaca dataset~\citep{alpaca,self-instruct} for three epochs. The loss curve of Adam during these three epochs exhibits a step-like decline, achieving a significantly smaller empirical loss compared to SGD.

Through our analysis on Adam above, we found that its second-order moment estimation has a significantly greater impact on its convergence than the first-order moment estimation. The second-order moment estimation is particularly effective for handling sparse data, allowing parameters that are infrequently updated to receive larger update steps.

Furthermore, the second-order moment in the optimizer's state has been proven to be decomposable or compressible to reduce memory usage. For example, Adafactor~\cite{adafactor} decomposes the second moment \( {v}_{t,i} \in \mathbb{R}^{m \times n} \) by minimizing the I-divergence into \( {r}_{t,i} \in \mathbb{R}^{m \times 1} \) and \( {c}_{t,i} \in \mathbb{R}^{1 \times n} \) such that 
\begin{gather}
    {v}_{t,i} = {r}_{t,i}{c}_{t,i}/(\bm{1}_m^T r_{t,i}).
\end{gather}

The update formulas for \( {r}_{t,i} \) and \( {c}_{t,i} \) in Adafactor are as follows:

\begin{align}
    {r}_{t,i} &= \beta_1 {r}_{t-1,i} + (1-\beta_1) {g}_{t,i}^2\ \bm{1}_n, \\
    {c}_{t,i} &= \beta_2 {c}_{t-1,i} + (1-\beta_2) \bm{1}_m^T {g}_{t,i}^2,
\end{align}
where \( \bm{1}_n \) and \( \bm{1}_m^T \) are all-ones vectors of dimensions \( n \times 1 \) and \( 1 \times m \), respectively.
\begin{table*}[t]
\small
\centering
\resizebox{0.9\textwidth}{!}{
\begin{tabular}{@{}llcccccc@{}}
\toprule
\textbf{Model}                      & \textbf{Method} & \textbf{MMLU} & \textbf{BBH}  & \textbf{GSM8K} & \textbf{HumanEval} & \textbf{AlpacaFarm} & \textbf{Avg.} \\ \midrule
\multirow{5}{*}{LLaMA-7B}  & N/A        & 31.5 & 32.3 & 10.9  & 11.6      & 4.2        & 18.1 \\
                           & LoRA            & 33.5 & 34.8 & 12.3  & 11.0      & 41.1       & 26.5 \\
                           & AdamW           & 39.3 & 34.4 & 9.6   & 11.6      & 50.6       & 29.1 \\
                           & LOMO            & 30.7 & 34.0 & 12.0  & \textbf{12.8}      & 30.6       & 24.0 \\
                           \rowcolor[gray]{0.9} \cellcolor{white}
                           & AdaLomo         & \textbf{39.5} & \textbf{36.0} & \textbf{14.4 } & 11.0      & \textbf{53.3}      & \textbf{30.8} \\ \midrule
\multirow{5}{*}{LLaMA-13B} & N/A        & 45.2 & 38.5 & 19.5  & 14.0      & 5.3        & 24.5 \\
                           & LoRA            & 48.3 & 40.3 & 20.2  & \textbf{19.5}      & 49.1       & 35.5 \\
                           & AdamW           & 49.4 & 40.2 & 21.8  & 18.9      & 61.0       & 38.2 \\
                           & LOMO            & 44.2 & 38.9 & 21.3  & 16.5      & 38.4       & 31.8 \\
                           \rowcolor[gray]{0.9} \cellcolor{white}
                           & AdaLomo         & \textbf{50.0} & \textbf{41.5} & \textbf{25.3}  & 18.9      & \textbf{62.9}       & \textbf{39.7} \\ \midrule
\multirow{5}{*}{LLaMA-30B} & N/A        & 57.7 & 51.8 & 40.3  & 20.1      & 7.1        & 35.4 \\
                           & LoRA            & 59.3 & \textbf{52.3} & 42.8  & \textbf{26.2}      & 63.3       & 48.8 \\
                           & AdamW           & 57.3 & 49.5 & 36.6  & 21.3      & 65.5       & 46.1 \\
                           & LOMO            & 56.3 & 51.5 & 44.4  & 18.9      & 57.8       & 45.8 \\
                           \rowcolor[gray]{0.9} \cellcolor{white}
                           & AdaLomo         & \textbf{59.4} & 52.1 & \textbf{48.5}  & 25.6      & \textbf{69.6}       & \textbf{51.0} \\ \midrule
\multirow{5}{*}{LLaMA-65B} & N/A        & 62.4 & 58.7 & 53.9  & 20.7      & 4.7        & 40.1 \\
                           & LoRA            & 62.7 & 58.7 & \textbf{60.5}  & \textbf{32.9}      & 69.6       & \textbf{56.9} \\
                           & AdamW           & \textbf{63.0} & 57.9 & 55.3  & 28.1      & 73.1       & 55.5 \\
                           & LOMO            & 62.1 & 56.9 & 57.6  & 28.1      & 65.2       & 54.0 \\
                           \rowcolor[gray]{0.9} \cellcolor{white} & AdaLomo         & 62.7 & \textbf{59.0} & 59.7  & 29.9      & \textbf{73.4}       & \textbf{56.9} \\ \bottomrule
\end{tabular}
}
\caption{Performance of the LLaMA series models on various benchmarks after instruction-tuning with different optimization techniques. Bolded numbers indicate the best results for models of the same size on a given benchmark. ``N/A" denotes that no instruction-tuning is performed.}
\label{tab:instruction-tuning}
\end{table*}

\section{Method}
In this section, we introduce our proposed memory-efficient optimization algorithm, Adalomo. This algorithm has demonstrated performance comparable to the current de facto optimization method for large language models, AdamW, requiring less memory consumption. 

\subsection{AdaLomo}
\begin{algorithm}[ht]
\caption{AdaLomo}\label{alg:adalomo}
\begin{algorithmic}[1]
\Require model $f(\cdot)$ with parameter $\boldsymbol{\theta}$ , learning rate $\alpha$, max step $T$, training dataset $\mathcal{D}$, loss function $\mathcal{L}$, decay coefficient $\beta$, regularization constant $\epsilon$
\vspace{0.1cm}
\For{$t = 1$ \textbf{to} $T$}
    \State sample batch $\mathcal{B}=(\boldsymbol{x},\boldsymbol{y}) \subset \mathcal{D}$
    \State $\boldsymbol{\hat{y}} \gets f(\boldsymbol{x},\boldsymbol{\theta})$ \Comment{forward pass}
    \State $\ell \gets \mathcal{L}(\boldsymbol{y},\boldsymbol{\hat{y}})$
    
    \For{each parameter $\theta_i$ in the order of backpropagation}
        \State $g_{t,i}=\nabla_{\theta_{t-1,i}} \ell$ \Comment{$g_{t,i-1}$ needed for computing $g_{t,i}$}
        \State $r_{t,i} = \beta r_{t-1,i} + (1-\beta) g_{t,i}^2\bm{1}_n$
        \State $c_{t,i} = \beta c_{t-1,i} + (1-\beta) \bm{1}_m^Tg_{t,i}^2$
        \State $v_{t,i} = r_{t,i}c_{t,i}/(\bm{1}_m^Tr_{t,i})$
        \State $u_{t,i} = g_{t,i} / v_{t,i}$
        \State $\hat{u}_{t,i}=u_{t,i}/\text{max}(1,RMS(u_{t,i}))\times \text{max}(\epsilon, RMS(\theta_{t-1,i}))$
        \State $\theta_{t,i}=\theta_{t-1,i}-\alpha_t \hat{u}_{t,i}$
        \State ${g_{t,i-1}} \gets$ None \Comment{clear ${g_{t,i-1}}$}
    \EndFor
\EndFor
\end{algorithmic}
\end{algorithm}

Based on the analysis in Section~\ref{subsec:anaysis_moments}, to achieve improved optimization while maintaining low memory consumption, we decided to incorporate a second-order moment estimation and discard the first-order moment. In our pursuit of further memory efficiency, we applied non-negative matrix factorization to the second-order moment, inspired by Adafactor. 
Specifically, for each parameter \( \theta_i \) within the model parameters \( \bm{\theta} \), we introduce two optimizer states, \( r_i \) and \( c_i \). For parameters of size \( m \times n \), we store only \( r_i \) and \( c_i \) instead of storing \( v_i \). The size of the optimizer states is \( m + n \), which is negligible compared to the size of the parameters. 

Contrary to Adafactor, we update the optimizer state, update the parameters and discard the gradients during the gradient backpropagation process, which reduces our memory footprint to just 40\% of that required by Adafactor.
During parameter updates, we compute \( v_i=r_i c_i\) using \(r_i\) and \(c_i\) to provide adaptive learning rate for the parameters, which ensures that the optimization of AdaLomo is theoretically superior to that of LOMO based on the preceding analysis.
Additionally, we employ grouped update normalization, which nearly doubles training speed compared to the naive gradient norm used in LOMO.
The details of the algorithm are presented in Algorithm~\ref{alg:adalomo}.

\subsection{Grouped Update Normalization}
We utilize grouped update normalization in the AdaLomo update process, which entails adaptive modifications for the update of each parameter and helps maintain model stability especially during large-scale training. 
Grouped update normalization ensures that each parameter's update is meaningful and not overshadowed by large gradient values from other parameters, facilitating faster convergence and sustained stability.
In contrast, global update normalization, where all parameters share a single scaling factor, might lead to some parameters updating too rapidly or too slowly, thereby affecting both convergence speed and stability. This is especially evident in large language models where different layers and parameters can exhibit considerable variations in gradient magnitudes, rendering global scaling potentially less effective.

As shown in line 11 of Algorithm~\ref{alg:adalomo}, for the update matrix \( u_i \) for parameter \(\theta_i \), before applying it to the weight matrix, we divide it by the parameter-wise root-mean-square (RMS) of \( u_i \)
\footnote{The root-mean-square (RMS) of \( u \) is given by \( RMS(u) = \sqrt{\frac{\sum _{i=1}^{n}u_i^2}{n}} \), where \( n \) is the number of elements in \( u \).}. 
Additionally, we utilize the parameter-wise RMS of \( \theta_i \) to ensure the update step size is proportional to the magnitude of the parameter.

Furthermore, it's worth noting that grouped update normalization integrates seamlessly with AdaLomo's fused backward process. While global update normalization requires two backward passes as gradient normalization mentioned in Section~\ref{sec:grad_norm}, grouped update normalization allows us to normalize the update matrices within a single fused backward pass.

\begin{figure*}[ht]
    \centering
    \begin{subfigure}{0.46\textwidth}
        \centering
        \includegraphics[width=\textwidth]{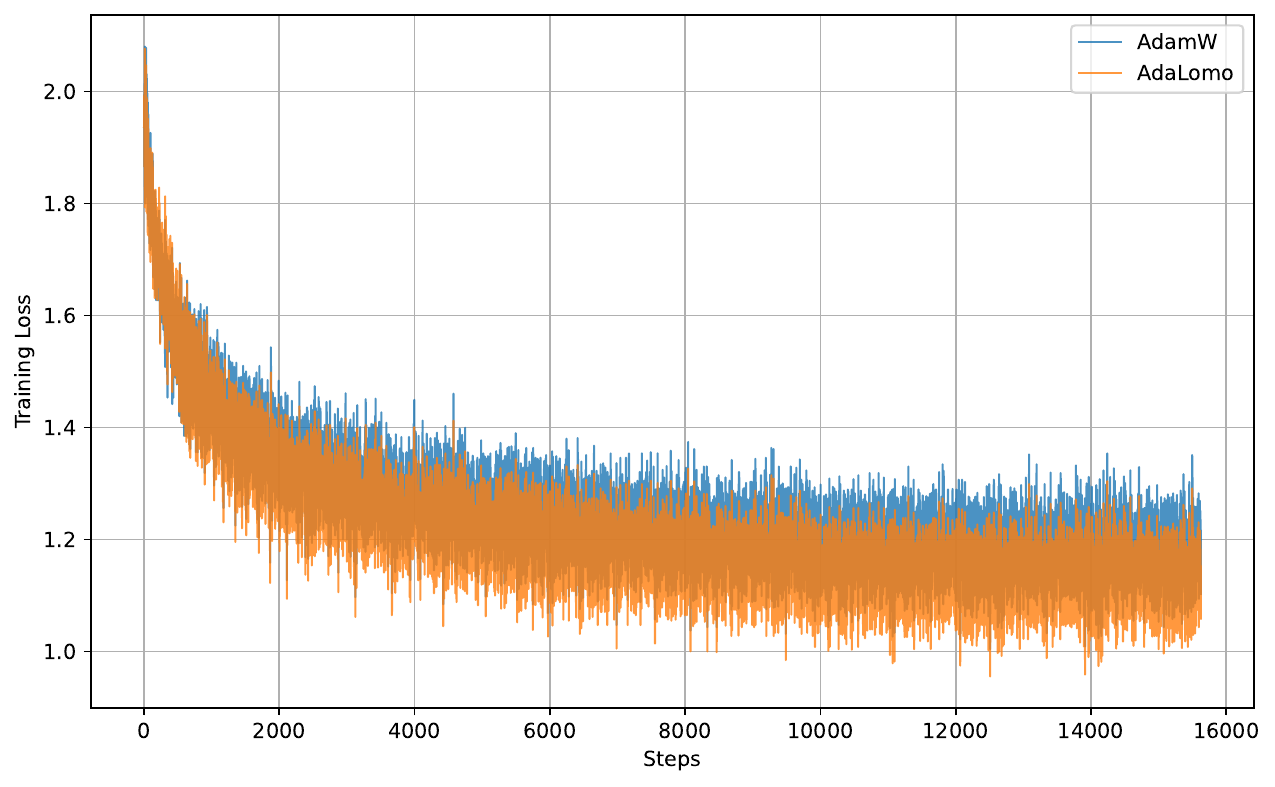}
        \caption{Training loss curve for LLaMA-7B.}
        \label{subfig:cn_7b_loss}
    \end{subfigure}
    \hfill
    \begin{subfigure}{0.46\textwidth}
        \centering
        \includegraphics[width=\textwidth]{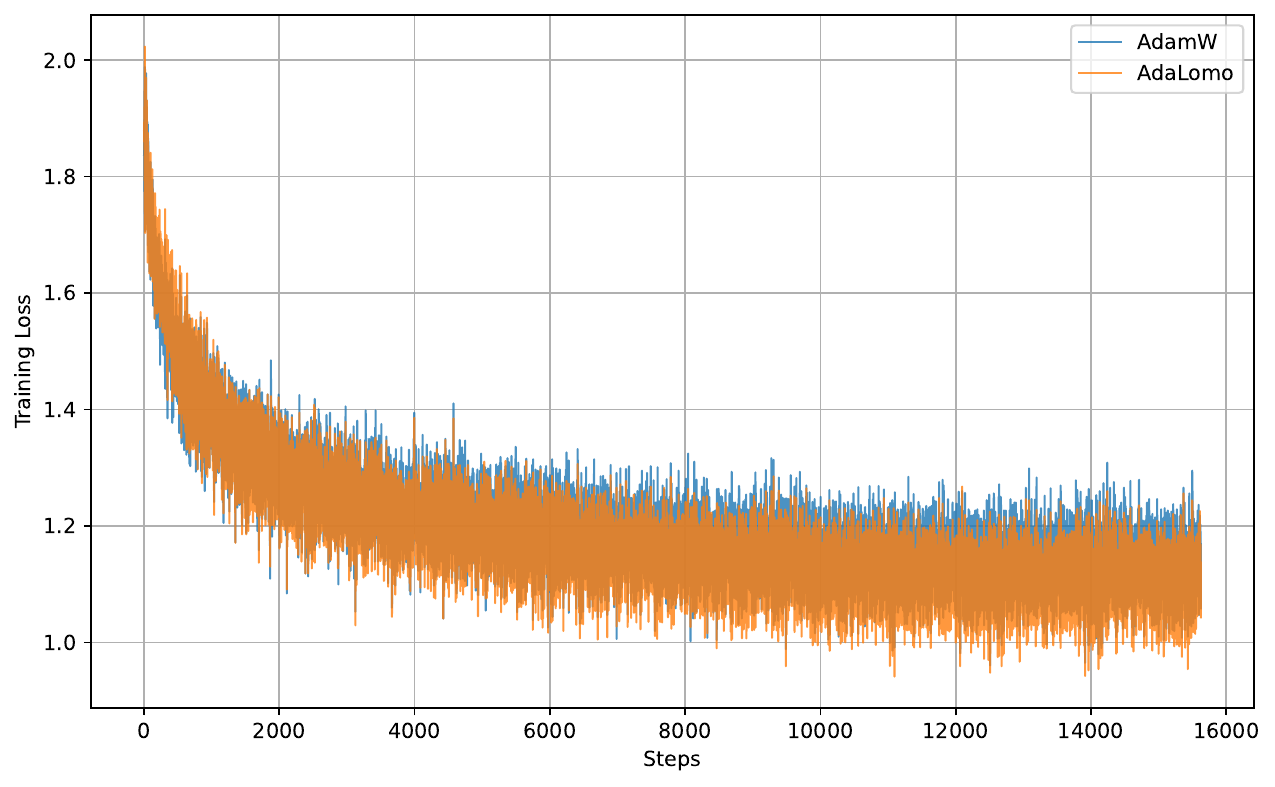}
        \caption{Training loss curve for LLaMA-13B.}
        \label{subfig:cn_13b_loss}
    \end{subfigure}
    
    \vspace{0.3em} 
    
    \begin{subfigure}{0.46\textwidth}
        \centering
        \includegraphics[width=\textwidth]{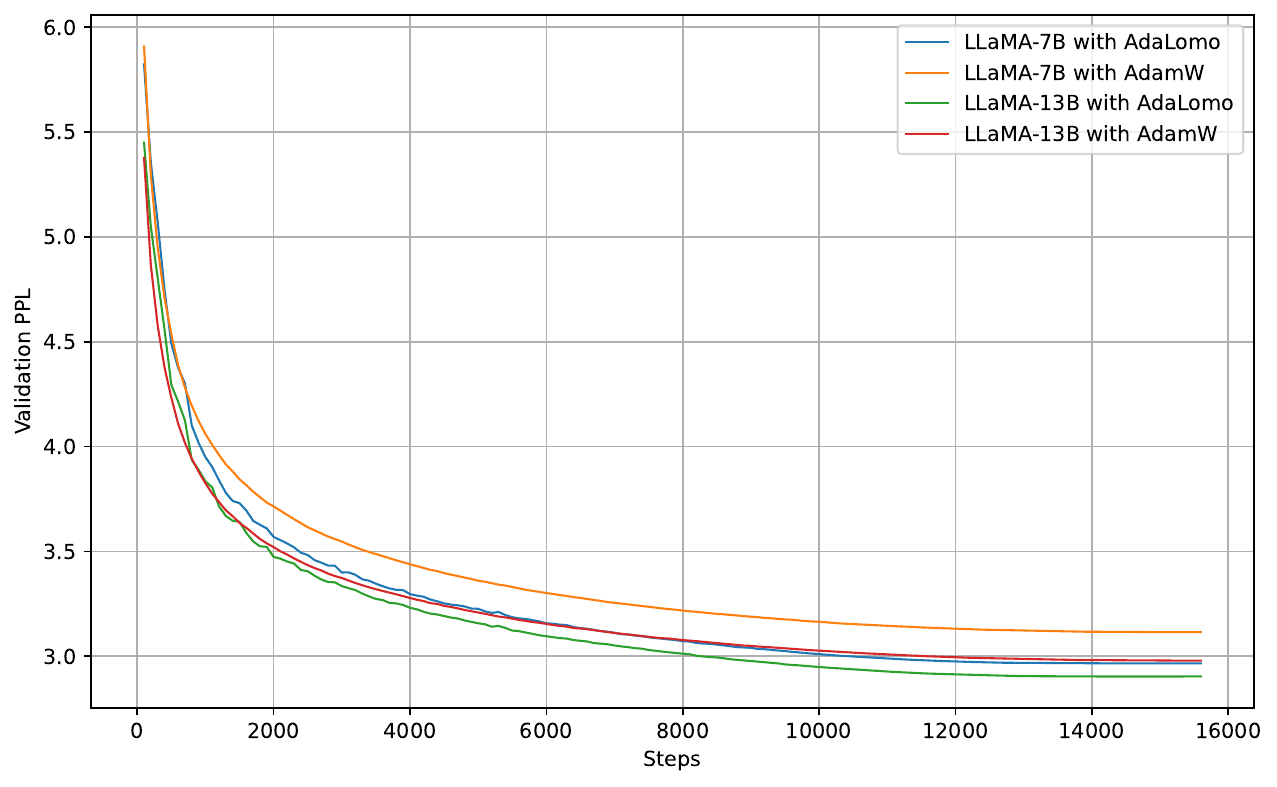}
        \caption{Perplexity of the validation set.}
        \label{subfig:cn_ppl}
    \end{subfigure}
    \hfill
    \begin{subfigure}{0.46\textwidth}
        \centering
        \includegraphics[width=\textwidth]{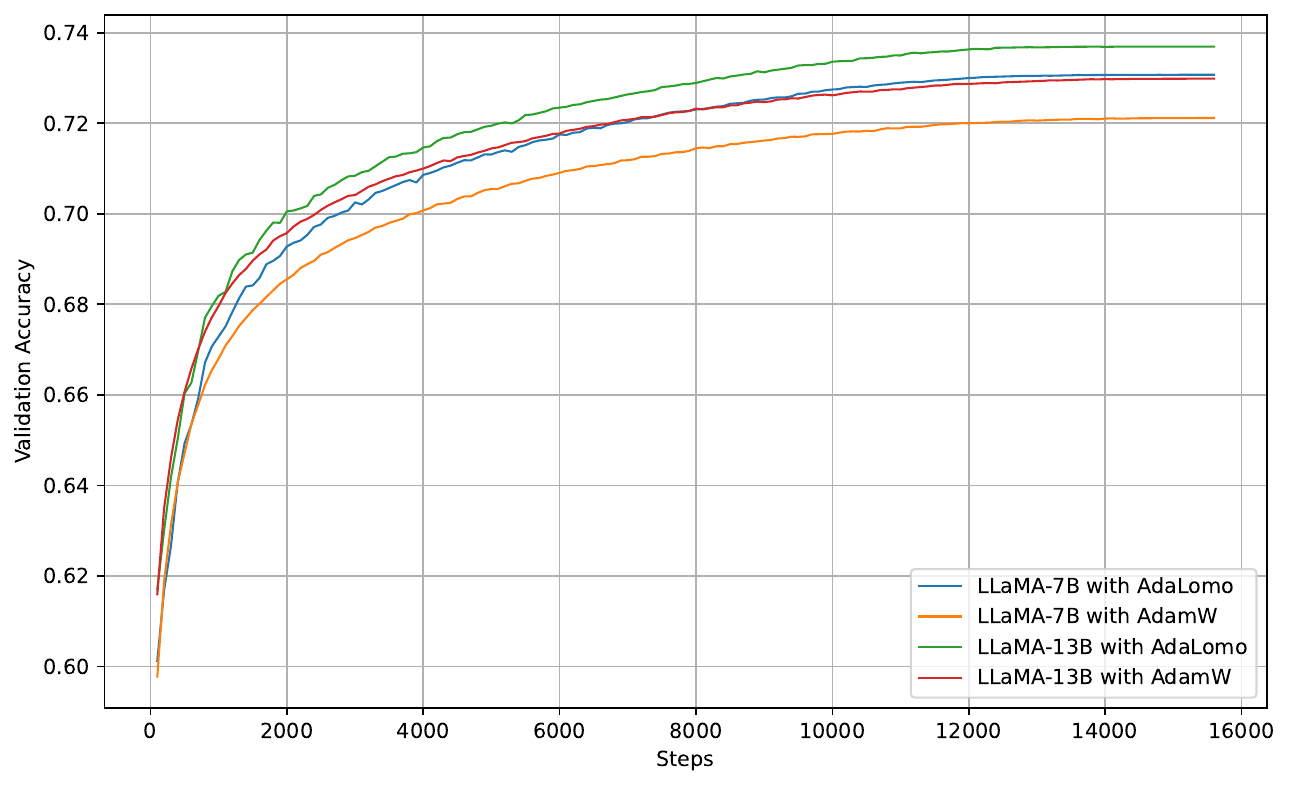}
        \caption{Next-token accuracy of the validation set.}
        \label{subfig:cn_acc}
    \end{subfigure}
    \caption{Results of further pre-training in the Chinese domain.}
\end{figure*}
\section{Experiments}

In this section, we evaluate the efficacy of AdaLomo in instruction-tuning, further pre-training and from-scratch pre-training. Additionally, we assess memory usage and throughput. Experiments are performed using the LLaMA series of models, which have parameter sizes ranging from 7 billion to 65 billion.

\subsection{Instruction Tuning}

We utilized GPT-4-Alpaca~\citep{gpt4-alpaca} as the training data to fine-tune LLaMA, incorporating 52k instruction-following demonstrations generated by GPT-4 using the Alpaca method. Besides the unaltered vanilla model and LOMO, we compared LoRA and AdamW, two prevalent methods for instruction-tuning large language models, which act as strong baselines.

We evaluated the trained models across diverse tasks: knowledge-based tasks (MMLU~\citep{mmlu}), general reasoning tasks (BBH~\citep{bbh}), mathematical tasks (GSM8K~\citep{gsm8k}), coding tasks (HumanEval~\citep{humaneval}), and instruction-following tasks (AlpacaFarm~\citep{alpaca-farm}). For MMLU, BBH, and GSM8K, the answers are obtained by generating, and are assessed using accuracy. The HumanEval task is evaluated using pass@1. The AlpacaFarm task is assessed by comparing the win rate of responses against those from GPT-3.5~\citep{gpt3}, as scored by GPT-4~\citep{gpt4}. Training and evaluation are conducted using templates provided in the Alpaca repository. 

The results are presented in Table~\ref{tab:instruction-tuning}. 
Compared to the vanilla model, models trained using these methods generally exhibit improved performance, especially in instruction-following capabilities. LOMO's performance on general reasoning (BBH), mathematics (GSM8K), and coding (HumanEval) tasks was comparable to that of LoRA and AdamW across all model sizes. However, its performance on knowledge-based tasks (MMLU) and instruction-following tasks (AlpacaFarm) is relatively inferior. The performance gap between LOMO and both LoRA and AdamW on these two tasks decreases as the model size increases.
By incorporating the second-order moment estimation, AdaLomo addresses LOMO's limitations, achieving comparable results with AdamW across various benchmarks for all model sizes.

\subsection{Further Pre-training}

\begin{figure*}[ht]
    \centering
    \begin{subfigure}{0.46\textwidth}
        \centering
        \includegraphics[width=\textwidth]{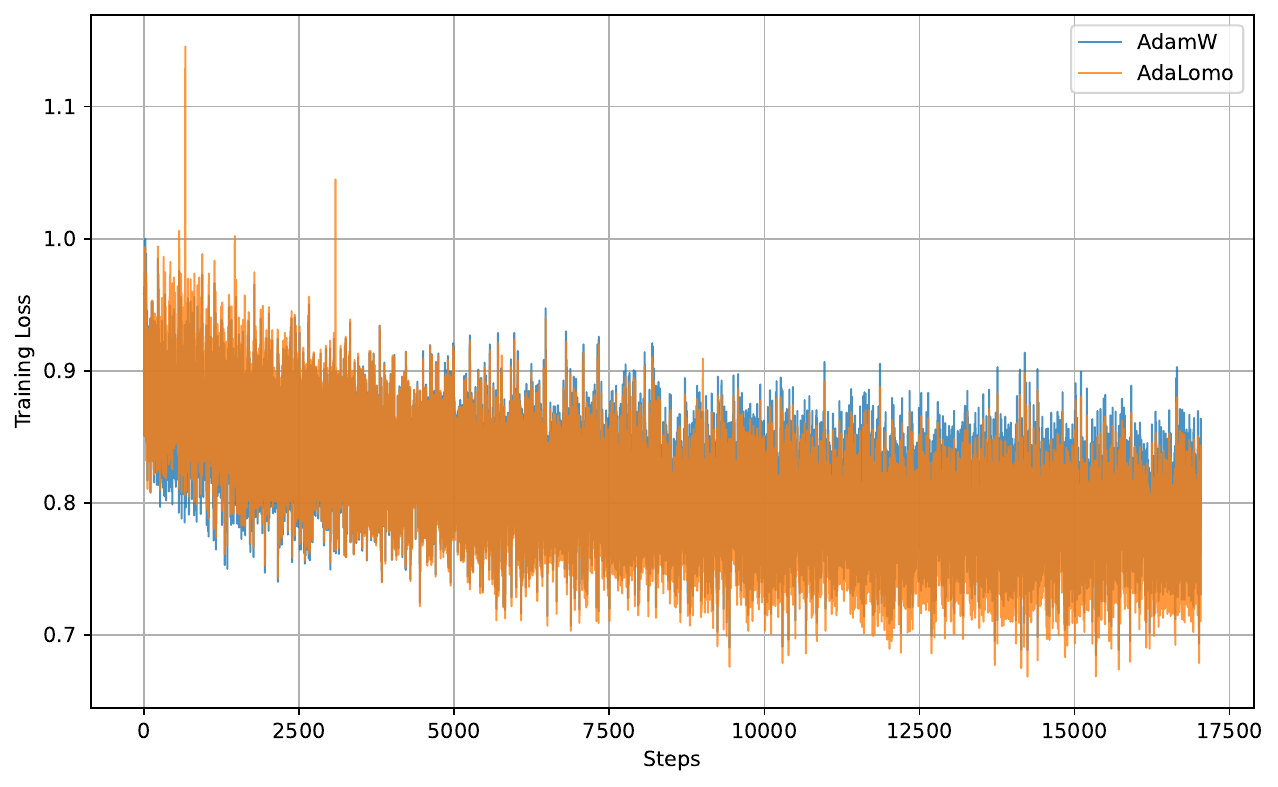}
        \caption{Training loss curve for LLaMA-7B.}
        \label{subfig:py_7b_loss}
    \end{subfigure}
    \hfill
    \begin{subfigure}{0.46\textwidth}
        \centering
        \includegraphics[width=\textwidth]{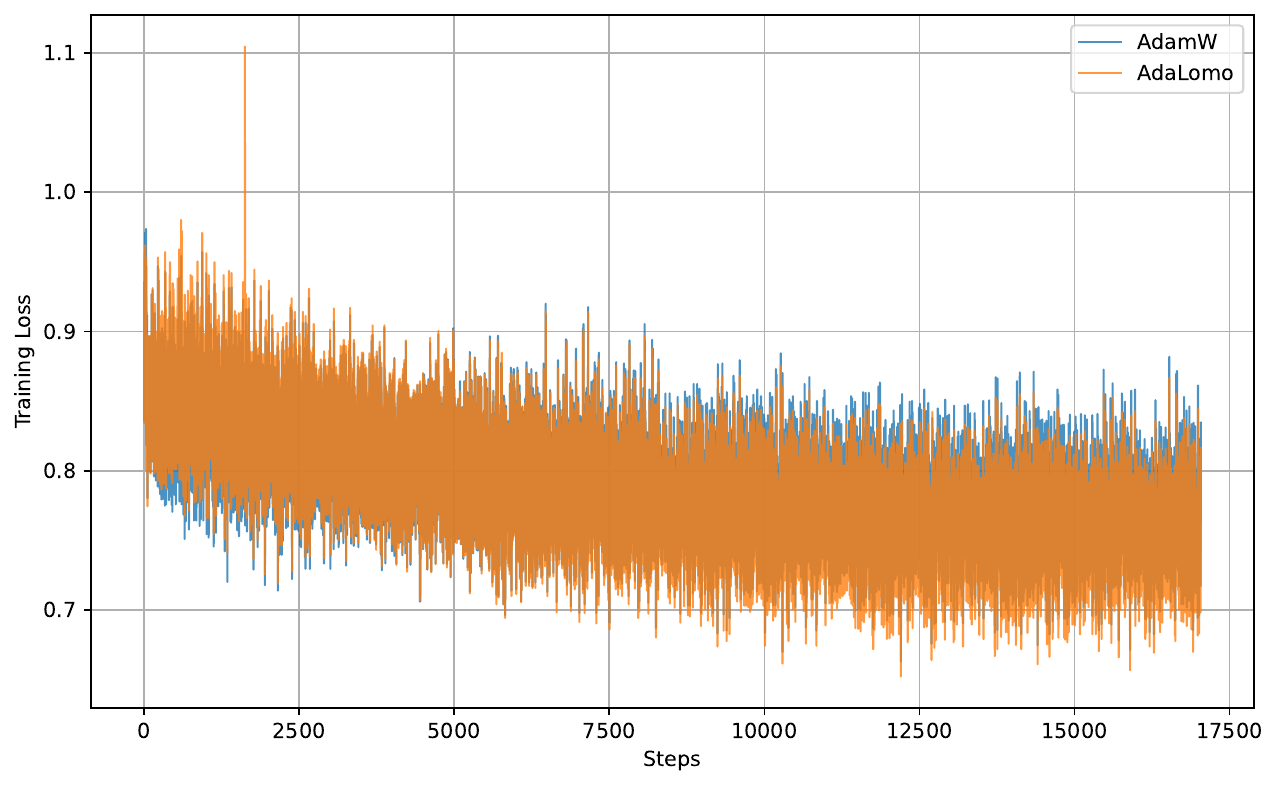}
        \caption{Training loss curve for LLaMA-13B.}
        \label{subfig:py_13b_loss}
    \end{subfigure}
    
    \vspace{0.3em} 
    
    \begin{subfigure}{0.46\textwidth}
        \centering
        \includegraphics[width=\textwidth]{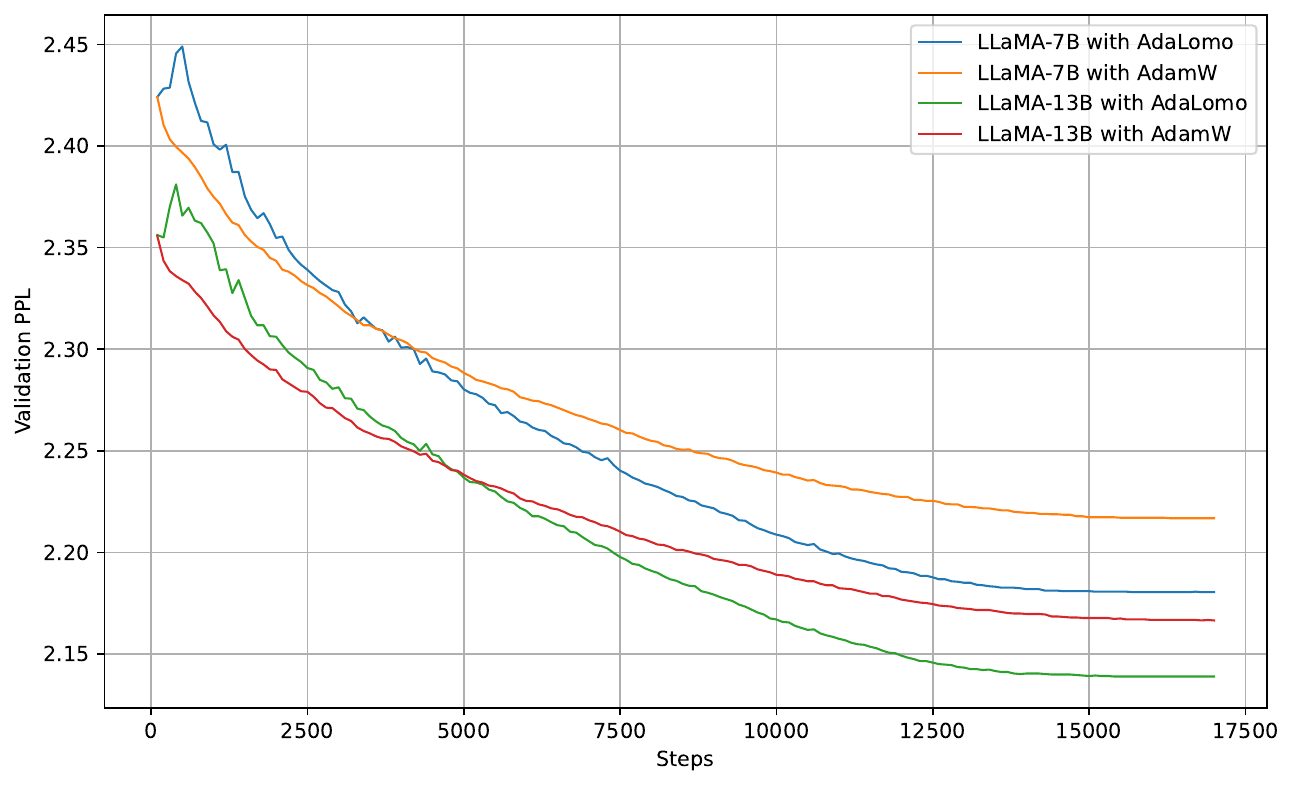}
        \caption{Perplexity of the validation set.}
        \label{subfig:py_ppl}
    \end{subfigure}
    \hfill
    \begin{subfigure}{0.46\textwidth}
        \centering
        \includegraphics[width=\textwidth]{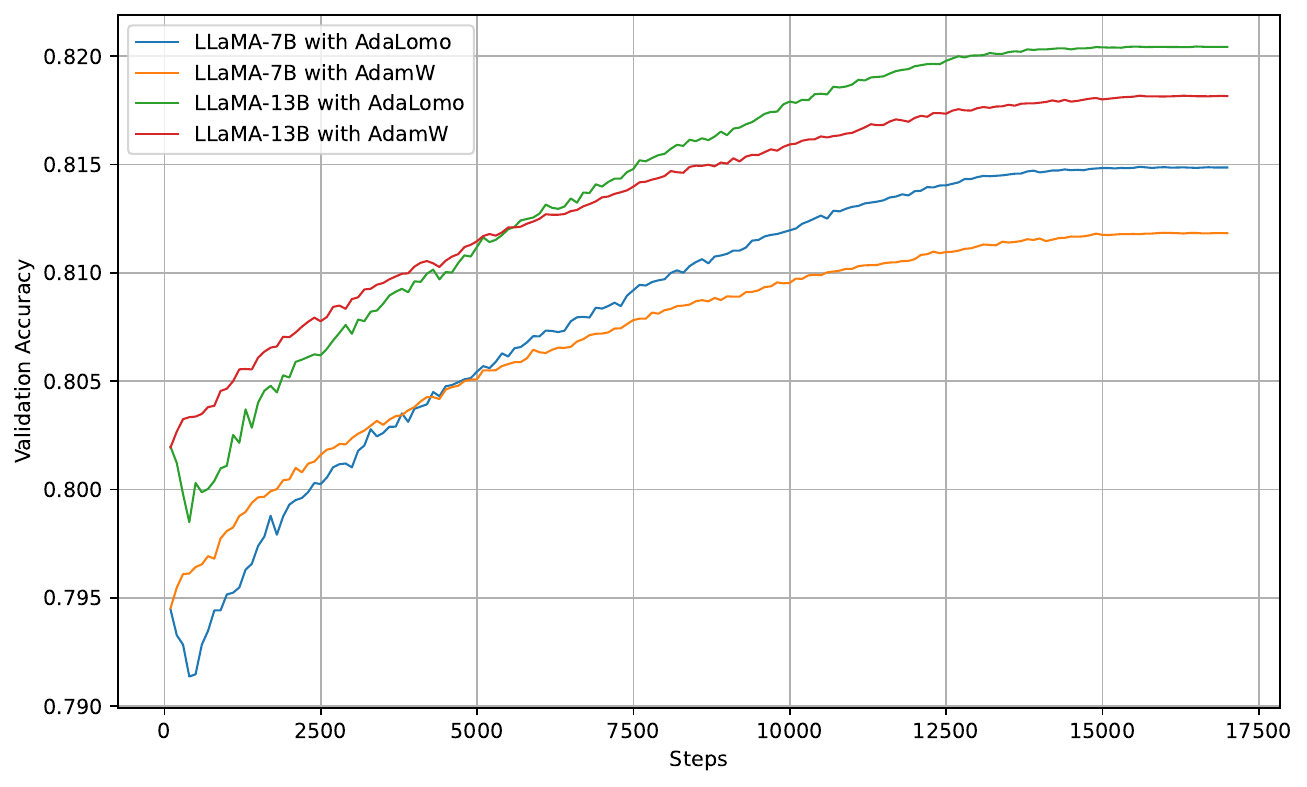}
        \caption{Next-token accuracy of the validation set.}
        \label{subfig:py_acc}
    \end{subfigure}
    \caption{Results of further pre-training in the Python code domain.}
    \label{fig:py}
\end{figure*}

Further pre-training refers to the additional large-scale unsupervised learning applied to a pre-trained model. 
We conduct further pre-training on the LLaMA model with parameter sizes of 7B and 13B in two domains: Chinese and Python code. The LLaMA model had limited exposure to data from these two domains during its initial pre-training phase.
Baidu-baike is a Chinese online encyclopedia. We scraped 2 million entries from Baidu-baike for further pre-training in the Chinese domain. Additionally, we extracted 2.2 million entries from the Python subset of the StarCoder~\citep{starcoder} training dataset for further pre-training in the Python code domain. Beyond this, we set aside 2,000 entries as a validation set. 

We choose AdamW as the baseline for comparison. The training hyper-parameters and data samples are detailed in Appendix~\ref{appendix:further-train}. We plot the loss curve during the model's training process and tested the perplexity and accuracy of the next-token prediction every 100 steps on the validation set.

As shown in Figure~\ref{subfig:cn_7b_loss} and \ref{subfig:cn_13b_loss}, during the further pre-training in Chinese, the loss curves of AdaLomo and AdamW overlap significantly, with AdaLomo's curve slightly below that of AdamW. The fluctuation range of their losses is at a similar level. Figure~\ref{subfig:cn_ppl} and \ref{subfig:cn_acc} also indicate that AdaLomo ultimately achieved a slightly lower perplexity and accuracy on the validation set than AdamW. Both methods effectively reduced LLaMA's perplexity in Chinese.

Figure~\ref{fig:py} presents the results of further pre-training in the Python code domain. The overall findings are similar to those in the Chinese domain, with some differences. 
 The enhancement of LLaMA's capabilities in the Python code domain through further pre-training is relatively less pronounced. 
This is because, in terms of perplexity, the original LLaMA performs better on Python code than on Chinese. 
Although AdaLomo exhibited some fluctuations during the initial warmup phase (with a perplexity difference of less than 0.02), it converged to a more optimal point at a faster rate thereafter. The LLaMA-13B model exhibited less fluctuation than the LLaMA-7B model. We attribute these fluctuations to AdaLomo's reliance on \(\bm{g}_t^2\) over \(\bm{v}_{t-1}\) during the early stages of training, and the fact that AdaLomo does not utilize momentum.

Grouped update normalization effectively substitutes the role of gradient normalization.
It enables stable training even without the use of gradient normalization, which is essential to prevent gradient explosion but with a decrease in throughput for LOMO. A detailed comparison regarding gradient normalization are shown in Appendix~\ref{appendix:grad_norm}.

\subsection{Pre-training from Scratch}
\begin{figure*}
    \centering
    \begin{subfigure}{0.32\textwidth}
        \centering
        \includegraphics[height=0.71\textwidth]{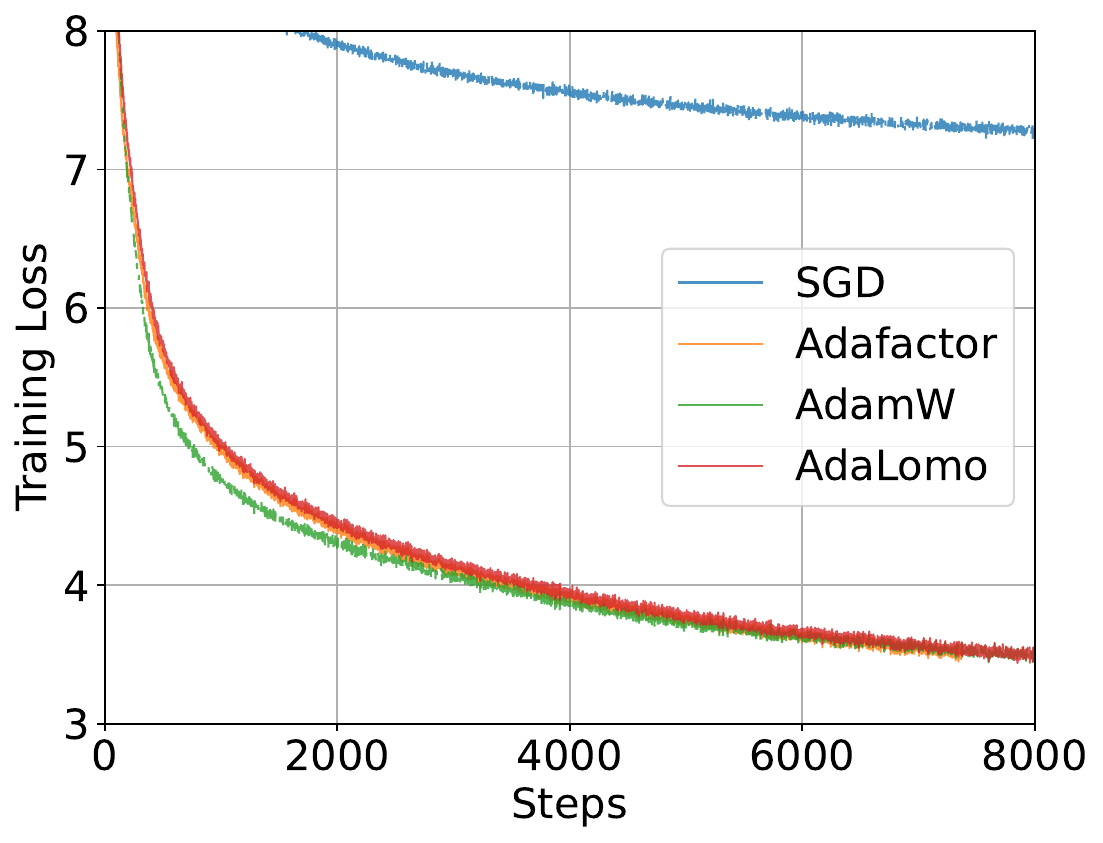}
        \caption{Training loss curve.}
    \end{subfigure}
    \hfill
    \begin{subfigure}{0.32\textwidth}
        \centering
        \includegraphics[height=0.71\textwidth]{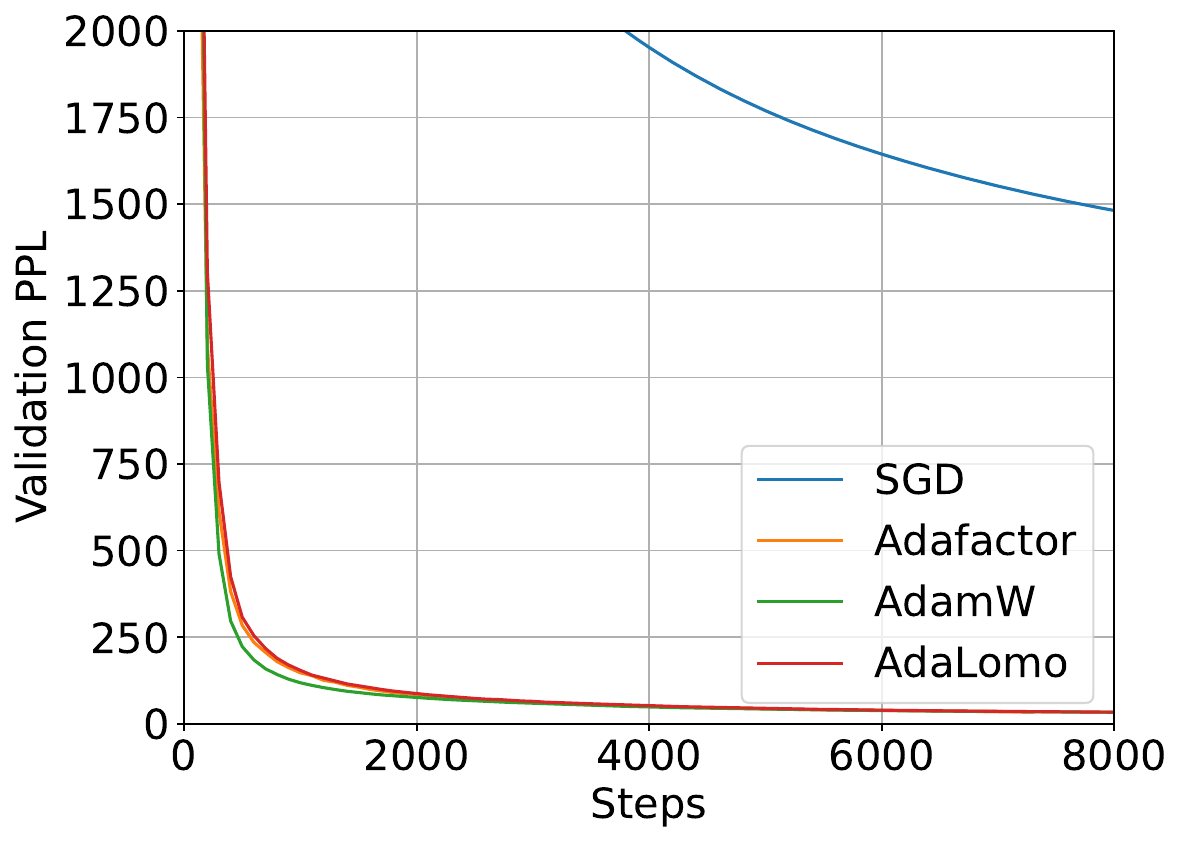}
        \caption{Validation perplexity.}
    \end{subfigure}
    \hfill
    \begin{subfigure}{0.32\textwidth}
        \centering
        \includegraphics[height=0.71\textwidth]{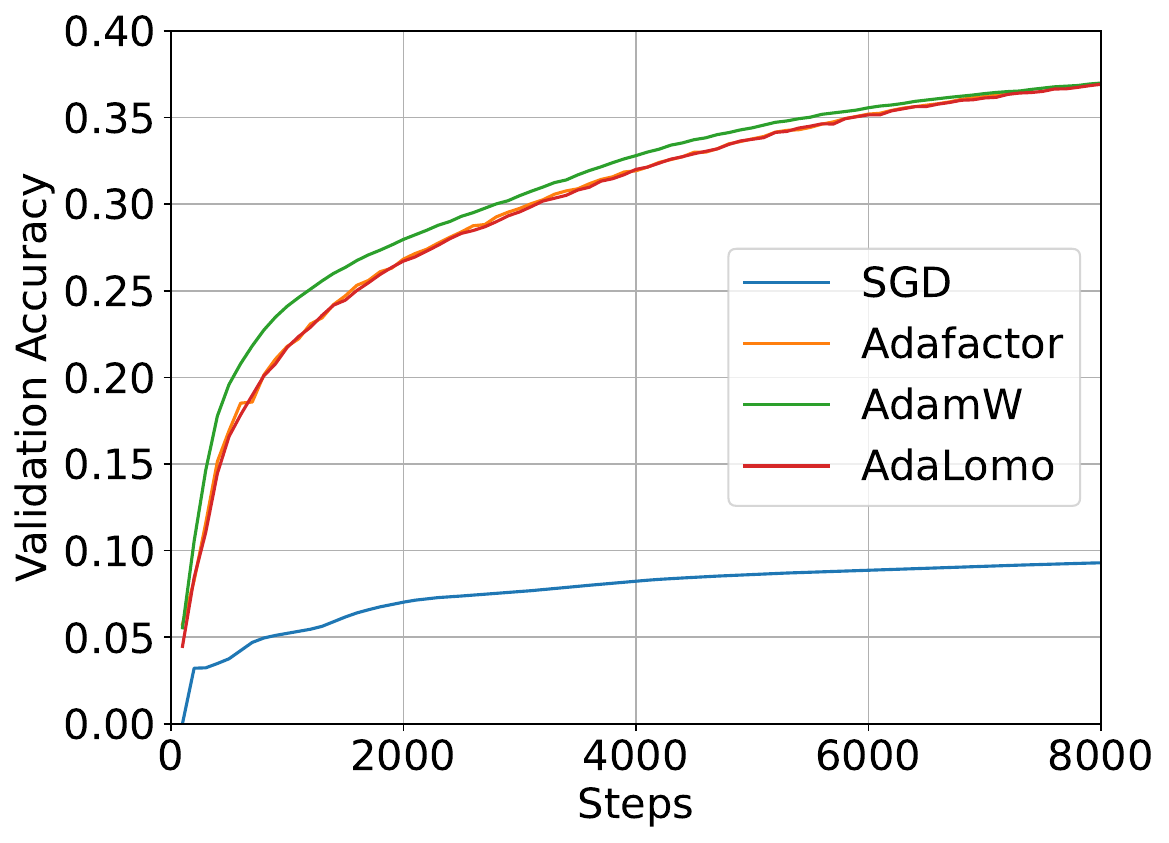}
        \caption{Validation next-token accuracy.}
    \end{subfigure}
    \caption{Results of pre-training LLaMA-1.1B from scratch on C4 corpus.}
    \label{fig:pretrain}
\end{figure*}

We conducted a from-scratch pre-training on the C4 corpus~\citep{t5c4} using a model with 1.1 billion parameters based on the LLaMA architecture\footnote{The architecture is consistent with TinyLlama-1.1B~\citep{zhang2024tinyllama}.}. The batch size was set to 1024, with a maximum data length of 2048 tokens, and warmup steps of 300 using a cosine scheduler. 
We report the training loss for the first 8000 steps, along with the perplexity and accuracy on the validation set, as shown in Figure~\ref{fig:pretrain}.

Our results indicate that AdamW, Adafactor, and AdaLomo exhibit comparable convergence performance, significantly outperforming SGD. This highlights the effectiveness of AdaLomo in the pretraining context.

\subsection{Memory and Throughput Profile}
\label{sec:profile}
\begin{figure*}[t]
    \centering
    \begin{subfigure}[b]{0.46\textwidth}
        \centering
        \includegraphics[width=\textwidth]{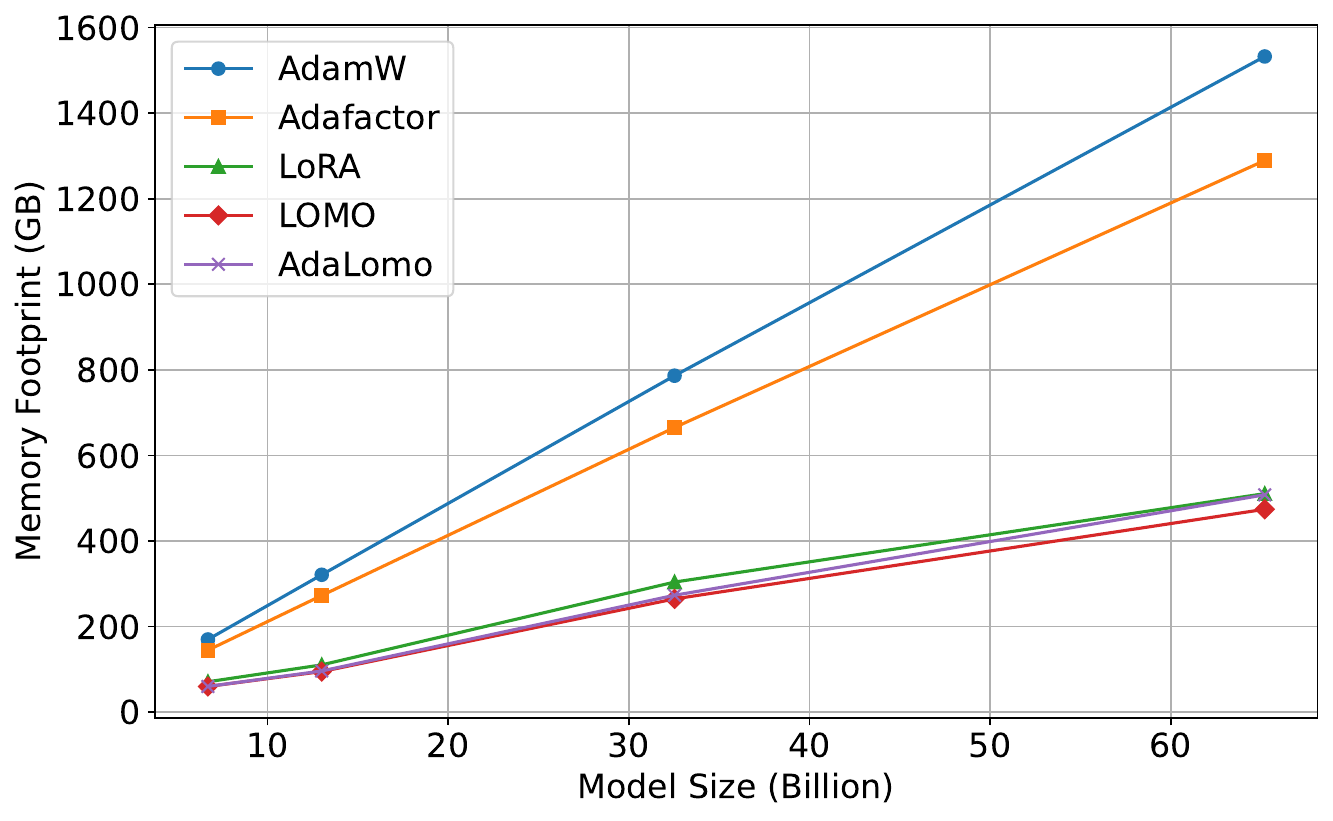}
        \caption{Memory usage with different methods.}
    \end{subfigure}
    \hfill
    \begin{subfigure}[b]{0.46\textwidth}
        \centering
        \includegraphics[width=\textwidth]{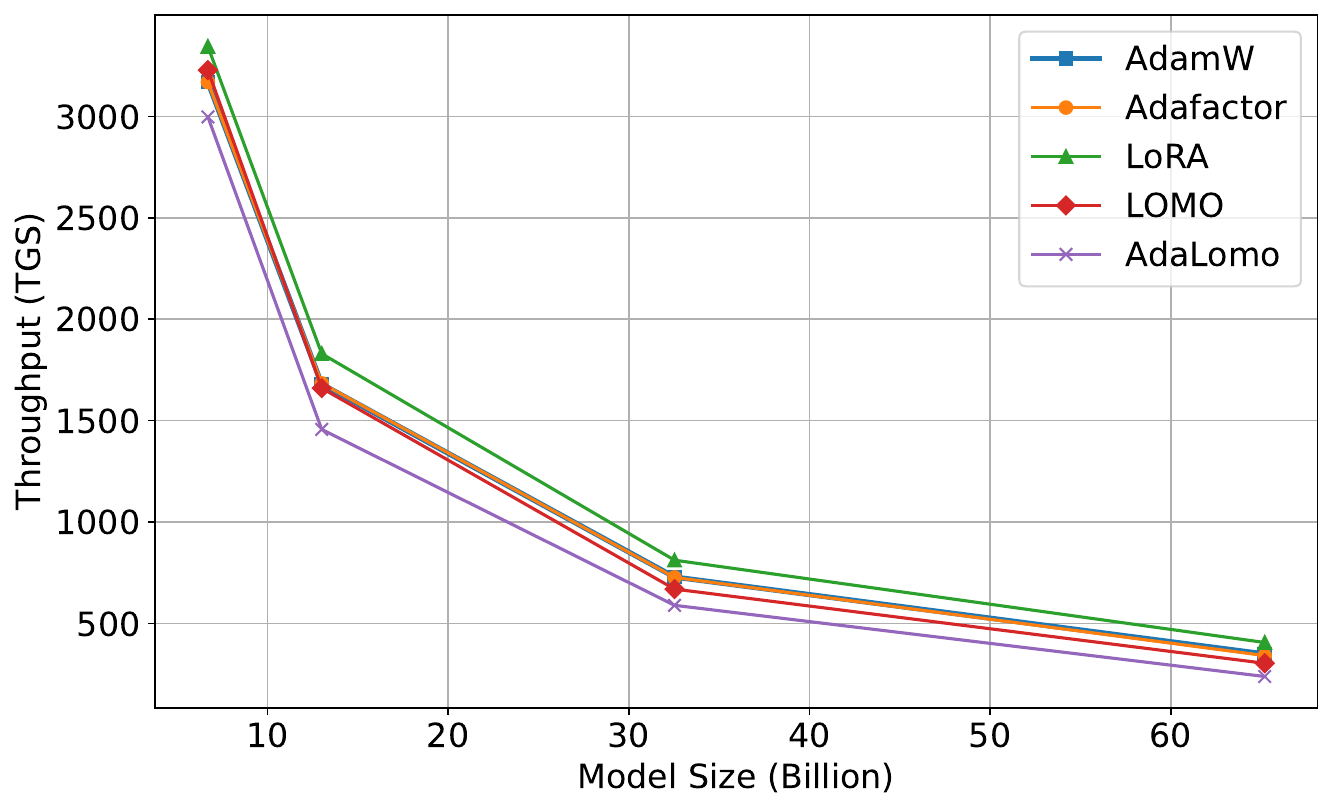}
        \caption{Throughput with different optimization methods.}
    \end{subfigure}
    \caption{Memory footprint and throughput using different optimization methods.}
    \label{fig:profile}
\end{figure*}
We evaluate the max allocated memory and throughput of AdamW, Adafactor, LoRA, LOMO, and AdaLomo, with the results in Figure~\ref{fig:profile}. We employ ZeRO-3~\citep{zero} for distributed training. 
Throughput is measured in terms of tokens processed per GPU per second (TGS). 
Detailed numerical results and more specific experimental settings can be found in Appendix~\ref{appendix:profile}.

Among the evaluated methods, AdamW exhibits the highest memory consumption. Adafactor reduces memory usage compared to AdamW by decomposing the second-order moment, resulting in memory savings proportional to the model's parameter size. AdaLomo, in comparison to LOMO, introduce an adaptive learning rate for each parameter. Nevertheless, its memory consumption remains close to that of LOMO and is comparable to LoRA, which trains with very few parameters. Due to fewer trainable parameters requiring communication during training, LoRA achieves the highest throughput. AdaLomo, which necessitates additional computations during parameter updates, shows slightly lower throughput than LOMO. 
All methods are tested with a consistent batch size, yet AdaLomo retains residual memory capacity, suggesting the potential for an increased batch size and greater throughput. 
Overall, the throughput of these methods is at the same level.

\section{Related Work}

Previous research has extensively explored memory-efficient optimizers. Adafactor~\citep{adafactor} employs non-negative matrix factorization and approximates the second-order moment estimate \( \bm{v} \in \mathbb{R}^{m\times n} \) using the outer product of \( \bm{r} \in \mathbb{R}^{m\times 1} \) and \( \bm{c} \in \mathbb{R}^{1 \times n} \), achieving sublinear memory consumption. The SM3 algorithm~\citep{sm3} introduces the cover of the parameters or, more specifically, a set of \( k \) non-empty parameter groups. Each parameter is assigned an adaptive learning rate based on this cover. 
For a parameter matrix of size \( m\times n \), the sets can be divided by rows and columns, resulting in \( m+n \) sets. This reduces the memory requirement from \( O(m\times n) \) to \( O(m+n) \), analogous to Adafactor's memory consumption. Another line to reduce memory usage is by utilizing low-precision storage for the optimizer state. \citet{dalle} and \citet{gopher} explored the stability of 16-bit optimizers. The 8-bit Optimizer~\citep{bit8optim}, using block-wise and dynamic exponent quantization, quantizes the optimizer states of SGDM and Adam to 8 bits. The 4-bit optimizer~\citep{bit4optim}, employing the newly proposed FP4 format and the adaptive gradient scaling technique.
To decrease the memory used by gradients, LOMO updates parameters simultaneously during the gradient computation in the backward pass.

Additionally, there exists a series of memory-efficient optimization methods designed exclusively for fine-tuning. BBT~\citep{bbt} and BBTv2~\citep{bbtv2} utilize evolutionary gradient-free algorithms to optimize continuous prompts without model updates. MeZO~\citep{mezo} employs zeroth-order optimization methods, estimating gradients using two forward passes and optimizing the model in-place, thus equating memory consumption with inference. Parameter-efficient fine-tuning~(PEFT)~\citep{peft_overview} methods selectively add or pick a subset of parameters for optimization, freezing the majority of the model parameters. 
In comparison, AdaLomo updates all parameters using a gradient-based method, suitable for both pre-training and fine-tuning, with memory consumption comparable to PEFT methods.

\section{Conclusion}
In this paper, we introduce AdaLomo, designed to reduce the training barriers for large language models. By incorporating an adaptive learning rate and utilizing grouped update normalization, AdaLomo achieves results comparable to AdamW in instruction-tuning, further pre-training and from-scratch pre-training. Concurrently, the memory footprint of AdaLomo is on par with the PEFT methods.

\section*{Limitations}
While AdaLomo is memory-efficient when training large language models, it primarily reduces the memory occupied by gradients and the optimizer states. Therefore, for models with a significant amount of activation values occupying memory, the reduction in memory usage by employing AdaLomo is limited. 
Thus, AdaLomo is best suited for training models with a large number of parameters. 
Additionally, while our experiments show that the throughput decrease is minimal, AdaLomo introduce some extra computational overhead, suggesting a direction for further improvement. This framework can be extended to optimizers using other update methods, such as SM3, and can also be adapted to methods related to optimizer states compression.

\section*{Ethics statement}
This paper employs open-source models LLaMA and the OpenAI API, all in compliance with their respective licenses. The datasets utilized, including AlpacaGPT4, MMLU, BBH, GSM8K, HumanEval and AlpacaFarm, permit public and freeusage. Resources used in constructing further pre-training datasets are openly available.

\section*{Acknowledgments}
This work was supported by the National Key Research and Development Program of China (No.2022ZD0160102). The computations in this research were performed using the CFFF platform of Fudan University.

\bibliography{custom}

\begin{thebibliography}{44}
\expandafter\ifx\csname natexlab\endcsname\relax\def\natexlab#1{#1}\fi

\bibitem[{Anil et~al.(2019)Anil, Gupta, Koren, and Singer}]{sm3}
Rohan Anil, Vineet Gupta, Tomer Koren, and Yoram Singer. 2019.
\newblock \href {https://proceedings.neurips.cc/paper/2019/hash/8f1fa0193ca2b5d2fa0695827d8270e9-Abstract.html} {Memory efficient adaptive optimization}.
\newblock In \emph{Advances in Neural Information Processing Systems 32: Annual Conference on Neural Information Processing Systems 2019, NeurIPS 2019, December 8-14, 2019, Vancouver, BC, Canada}, pages 9746--9755.

\bibitem[{Brown et~al.(2020)Brown, Mann, Ryder, Subbiah, Kaplan, Dhariwal, Neelakantan, Shyam, Sastry, Askell, Agarwal, Herbert{-}Voss, Krueger, Henighan, Child, Ramesh, Ziegler, Wu, Winter, Hesse, Chen, Sigler, Litwin, Gray, Chess, Clark, Berner, McCandlish, Radford, Sutskever, and Amodei}]{gpt3}
Tom~B. Brown, Benjamin Mann, Nick Ryder, Melanie Subbiah, Jared Kaplan, Prafulla Dhariwal, Arvind Neelakantan, Pranav Shyam, Girish Sastry, Amanda Askell, Sandhini Agarwal, Ariel Herbert{-}Voss, Gretchen Krueger, Tom Henighan, Rewon Child, Aditya Ramesh, Daniel~M. Ziegler, Jeffrey Wu, Clemens Winter, Christopher Hesse, Mark Chen, Eric Sigler, Mateusz Litwin, Scott Gray, Benjamin Chess, Jack Clark, Christopher Berner, Sam McCandlish, Alec Radford, Ilya Sutskever, and Dario Amodei. 2020.
\newblock \href {https://proceedings.neurips.cc/paper/2020/hash/1457c0d6bfcb4967418bfb8ac142f64a-Abstract.html} {Language models are few-shot learners}.
\newblock In \emph{Advances in Neural Information Processing Systems 33: Annual Conference on Neural Information Processing Systems 2020, NeurIPS 2020, December 6-12, 2020, virtual}.

\bibitem[{Chen et~al.(2021)Chen, Tworek, Jun, Yuan, de~Oliveira~Pinto, Kaplan, Edwards, Burda, Joseph, Brockman, Ray, Puri, Krueger, Petrov, Khlaaf, Sastry, Mishkin, Chan, Gray, Ryder, Pavlov, Power, Kaiser, Bavarian, Winter, Tillet, Such, Cummings, Plappert, Chantzis, Barnes, Herbert{-}Voss, Guss, Nichol, Paino, Tezak, Tang, Babuschkin, Balaji, Jain, Saunders, Hesse, Carr, Leike, Achiam, Misra, Morikawa, Radford, Knight, Brundage, Murati, Mayer, Welinder, McGrew, Amodei, McCandlish, Sutskever, and Zaremba}]{humaneval}
Mark Chen, Jerry Tworek, Heewoo Jun, Qiming Yuan, Henrique~Pond{\'{e}} de~Oliveira~Pinto, Jared Kaplan, Harrison Edwards, Yuri Burda, Nicholas Joseph, Greg Brockman, Alex Ray, Raul Puri, Gretchen Krueger, Michael Petrov, Heidy Khlaaf, Girish Sastry, Pamela Mishkin, Brooke Chan, Scott Gray, Nick Ryder, Mikhail Pavlov, Alethea Power, Lukasz Kaiser, Mohammad Bavarian, Clemens Winter, Philippe Tillet, Felipe~Petroski Such, Dave Cummings, Matthias Plappert, Fotios Chantzis, Elizabeth Barnes, Ariel Herbert{-}Voss, William~Hebgen Guss, Alex Nichol, Alex Paino, Nikolas Tezak, Jie Tang, Igor Babuschkin, Suchir Balaji, Shantanu Jain, William Saunders, Christopher Hesse, Andrew~N. Carr, Jan Leike, Joshua Achiam, Vedant Misra, Evan Morikawa, Alec Radford, Matthew Knight, Miles Brundage, Mira Murati, Katie Mayer, Peter Welinder, Bob McGrew, Dario Amodei, Sam McCandlish, Ilya Sutskever, and Wojciech Zaremba. 2021.
\newblock \href {http://arxiv.org/abs/2107.03374} {Evaluating large language models trained on code}.
\newblock \emph{CoRR}, abs/2107.03374.

\bibitem[{Cobbe et~al.(2021)Cobbe, Kosaraju, Bavarian, Chen, Jun, Kaiser, Plappert, Tworek, Hilton, Nakano, Hesse, and Schulman}]{gsm8k}
Karl Cobbe, Vineet Kosaraju, Mohammad Bavarian, Mark Chen, Heewoo Jun, Lukasz Kaiser, Matthias Plappert, Jerry Tworek, Jacob Hilton, Reiichiro Nakano, Christopher Hesse, and John Schulman. 2021.
\newblock \href {http://arxiv.org/abs/2110.14168} {Training verifiers to solve math word problems}.
\newblock \emph{CoRR}, abs/2110.14168.

\bibitem[{Darken et~al.(1992)Darken, Chang, Moody et~al.}]{lr}
Christian Darken, Joseph Chang, John Moody, et~al. 1992.
\newblock Learning rate schedules for faster stochastic gradient search.
\newblock In \emph{Neural networks for signal processing}, volume~2, pages 3--12. Citeseer.

\bibitem[{Dauphin et~al.(2014)Dauphin, Pascanu, G{\"{u}}l{\c{c}}ehre, Cho, Ganguli, and Bengio}]{saddle_point}
Yann~N. Dauphin, Razvan Pascanu, {\c{C}}aglar G{\"{u}}l{\c{c}}ehre, KyungHyun Cho, Surya Ganguli, and Yoshua Bengio. 2014.
\newblock \href {https://proceedings.neurips.cc/paper/2014/hash/17e23e50bedc63b4095e3d8204ce063b-Abstract.html} {Identifying and attacking the saddle point problem in high-dimensional non-convex optimization}.
\newblock In \emph{Advances in Neural Information Processing Systems 27: Annual Conference on Neural Information Processing Systems 2014, December 8-13 2014, Montreal, Quebec, Canada}, pages 2933--2941.

\bibitem[{Dettmers et~al.(2022)Dettmers, Lewis, Shleifer, and Zettlemoyer}]{bit8optim}
Tim Dettmers, Mike Lewis, Sam Shleifer, and Luke Zettlemoyer. 2022.
\newblock \href {https://openreview.net/forum?id=shpkpVXzo3h} {8-bit optimizers via block-wise quantization}.
\newblock In \emph{The Tenth International Conference on Learning Representations, {ICLR} 2022, Virtual Event, April 25-29, 2022}. OpenReview.net.

\bibitem[{Ding et~al.(2023)Ding, Qin, Yang, Wei, Yang, Su, Hu, Chen, Chan, Chen, Yi, Zhao, Wang, Liu, Zheng, Chen, Liu, Tang, Li, and Sun}]{peft_overview}
Ning Ding, Yujia Qin, Guang Yang, Fuchao Wei, Zonghan Yang, Yusheng Su, Shengding Hu, Yulin Chen, Chi{-}Min Chan, Weize Chen, Jing Yi, Weilin Zhao, Xiaozhi Wang, Zhiyuan Liu, Hai{-}Tao Zheng, Jianfei Chen, Yang Liu, Jie Tang, Juanzi Li, and Maosong Sun. 2023.
\newblock \href {https://doi.org/10.1038/s42256-023-00626-4} {Parameter-efficient fine-tuning of large-scale pre-trained language models}.
\newblock \emph{Nat. Mac. Intell.}, 5(3):220--235.

\bibitem[{Dubois et~al.(2023)Dubois, Li, Taori, Zhang, Gulrajani, Ba, Guestrin, Liang, and Hashimoto}]{alpaca-farm}
Yann Dubois, Xuechen Li, Rohan Taori, Tianyi Zhang, Ishaan Gulrajani, Jimmy Ba, Carlos Guestrin, Percy Liang, and Tatsunori~B. Hashimoto. 2023.
\newblock \href {https://doi.org/10.48550/arXiv.2305.14387} {Alpacafarm: {A} simulation framework for methods that learn from human feedback}.
\newblock \emph{CoRR}, abs/2305.14387.

\bibitem[{Duchi et~al.(2011)Duchi, Hazan, and Singer}]{adagrad}
John~C. Duchi, Elad Hazan, and Yoram Singer. 2011.
\newblock \href {https://doi.org/10.5555/1953048.2021068} {Adaptive subgradient methods for online learning and stochastic optimization}.
\newblock \emph{J. Mach. Learn. Res.}, 12:2121--2159.

\bibitem[{Hendrycks et~al.(2021)Hendrycks, Burns, Basart, Zou, Mazeika, Song, and Steinhardt}]{mmlu}
Dan Hendrycks, Collin Burns, Steven Basart, Andy Zou, Mantas Mazeika, Dawn Song, and Jacob Steinhardt. 2021.
\newblock \href {https://openreview.net/forum?id=d7KBjmI3GmQ} {Measuring massive multitask language understanding}.
\newblock In \emph{9th International Conference on Learning Representations, {ICLR} 2021, Virtual Event, Austria, May 3-7, 2021}. OpenReview.net.

\bibitem[{Hu et~al.(2022)Hu, Shen, Wallis, Allen{-}Zhu, Li, Wang, Wang, and Chen}]{lora}
Edward~J. Hu, Yelong Shen, Phillip Wallis, Zeyuan Allen{-}Zhu, Yuanzhi Li, Shean Wang, Lu~Wang, and Weizhu Chen. 2022.
\newblock \href {https://openreview.net/forum?id=nZeVKeeFYf9} {Lora: Low-rank adaptation of large language models}.
\newblock In \emph{The Tenth International Conference on Learning Representations, {ICLR} 2022, Virtual Event, April 25-29, 2022}. OpenReview.net.

\bibitem[{Kim et~al.(2021)Kim, Papamakarios, and Mnih}]{lipschitz_sa}
Hyunjik Kim, George Papamakarios, and Andriy Mnih. 2021.
\newblock The lipschitz constant of self-attention.
\newblock In \emph{International Conference on Machine Learning}, pages 5562--5571. PMLR.

\bibitem[{Kingma and Ba(2015)}]{adam}
Diederik~P. Kingma and Jimmy Ba. 2015.
\newblock \href {http://arxiv.org/abs/1412.6980} {Adam: {A} method for stochastic optimization}.
\newblock In \emph{3rd International Conference on Learning Representations, {ICLR} 2015, San Diego, CA, USA, May 7-9, 2015, Conference Track Proceedings}.

\bibitem[{Li et~al.(2023)Li, Allal, Zi, Muennighoff, Kocetkov, Mou, Marone, Akiki, Li, Chim, Liu, Zheltonozhskii, Zhuo, Wang, Dehaene, Davaadorj, Lamy{-}Poirier, Monteiro, Shliazhko, Gontier, Meade, Zebaze, Yee, Umapathi, Zhu, Lipkin, Oblokulov, Wang, V, Stillerman, Patel, Abulkhanov, Zocca, Dey, Zhang, Moustafa{-}Fahmy, Bhattacharyya, Yu, Singh, Luccioni, Villegas, Kunakov, Zhdanov, Romero, Lee, Timor, Ding, Schlesinger, Schoelkopf, Ebert, Dao, Mishra, Gu, Robinson, Anderson, Dolan{-}Gavitt, Contractor, Reddy, Fried, Bahdanau, Jernite, Ferrandis, Hughes, Wolf, Guha, von Werra, and de~Vries}]{starcoder}
Raymond Li, Loubna~Ben Allal, Yangtian Zi, Niklas Muennighoff, Denis Kocetkov, Chenghao Mou, Marc Marone, Christopher Akiki, Jia Li, Jenny Chim, Qian Liu, Evgenii Zheltonozhskii, Terry~Yue Zhuo, Thomas Wang, Olivier Dehaene, Mishig Davaadorj, Joel Lamy{-}Poirier, Jo{\~{a}}o Monteiro, Oleh Shliazhko, Nicolas Gontier, Nicholas Meade, Armel Zebaze, Ming{-}Ho Yee, Logesh~Kumar Umapathi, Jian Zhu, Benjamin Lipkin, Muhtasham Oblokulov, Zhiruo Wang, Rudra~Murthy V, Jason Stillerman, Siva~Sankalp Patel, Dmitry Abulkhanov, Marco Zocca, Manan Dey, Zhihan Zhang, Nour Moustafa{-}Fahmy, Urvashi Bhattacharyya, Wenhao Yu, Swayam Singh, Sasha Luccioni, Paulo Villegas, Maxim Kunakov, Fedor Zhdanov, Manuel Romero, Tony Lee, Nadav Timor, Jennifer Ding, Claire Schlesinger, Hailey Schoelkopf, Jan Ebert, Tri Dao, Mayank Mishra, Alex Gu, Jennifer Robinson, Carolyn~Jane Anderson, Brendan Dolan{-}Gavitt, Danish Contractor, Siva Reddy, Daniel Fried, Dzmitry Bahdanau, Yacine Jernite, Carlos~Mu{\~{n}}oz Ferrandis, Sean Hughes, Thomas
  Wolf, Arjun Guha, Leandro von Werra, and Harm de~Vries. 2023.
\newblock \href {https://doi.org/10.48550/arXiv.2305.06161} {Starcoder: may the source be with you!}
\newblock \emph{CoRR}, abs/2305.06161.

\bibitem[{Loshchilov and Hutter(2019)}]{adamw}
Ilya Loshchilov and Frank Hutter. 2019.
\newblock \href {https://openreview.net/forum?id=Bkg6RiCqY7} {Decoupled weight decay regularization}.
\newblock In \emph{7th International Conference on Learning Representations, {ICLR} 2019, New Orleans, LA, USA, May 6-9, 2019}. OpenReview.net.

\bibitem[{Lv et~al.(2023)Lv, Yang, Liu, Gao, Guo, and Qiu}]{lomo}
Kai Lv, Yuqing Yang, Tengxiao Liu, Qinghui Gao, Qipeng Guo, and Xipeng Qiu. 2023.
\newblock \href {https://doi.org/10.48550/arXiv.2306.09782} {Full parameter fine-tuning for large language models with limited resources}.
\newblock \emph{CoRR}, abs/2306.09782.

\bibitem[{Malladi et~al.(2023)Malladi, Gao, Nichani, Damian, Lee, Chen, and Arora}]{mezo}
Sadhika Malladi, Tianyu Gao, Eshaan Nichani, Alex Damian, Jason~D. Lee, Danqi Chen, and Sanjeev Arora. 2023.
\newblock \href {https://doi.org/10.48550/arXiv.2305.17333} {Fine-tuning language models with just forward passes}.
\newblock \emph{CoRR}, abs/2305.17333.

\bibitem[{Nesterov(1983)}]{nesterov_mom}
Yurii Nesterov. 1983.
\newblock A method for unconstrained convex minimization problem with the rate of convergence o (1/k2).
\newblock In \emph{Dokl. Akad. Nauk. SSSR}, volume 269, page 543.

\bibitem[{OpenAI(2023)}]{gpt4}
OpenAI. 2023.
\newblock \href {https://doi.org/10.48550/arXiv.2303.08774} {{GPT-4} technical report}.
\newblock \emph{CoRR}, abs/2303.08774.

\bibitem[{Peng et~al.(2023)Peng, Li, He, Galley, and Gao}]{gpt4-alpaca}
Baolin Peng, Chunyuan Li, Pengcheng He, Michel Galley, and Jianfeng Gao. 2023.
\newblock \href {https://doi.org/10.48550/arXiv.2304.03277} {Instruction tuning with {GPT-4}}.
\newblock \emph{CoRR}, abs/2304.03277.

\bibitem[{Qi et~al.(2023)Qi, Wang, and Zhang}]{Lipschitz}
Xianbiao Qi, Jianan Wang, and Lei Zhang. 2023.
\newblock \href {http://arxiv.org/abs/2306.09338} {Understanding optimization of deep learning via jacobian matrix and lipschitz constant}.

\bibitem[{Qian(1999)}]{momentum}
Ning Qian. 1999.
\newblock \href {https://doi.org/10.1016/S0893-6080(98)00116-6} {On the momentum term in gradient descent learning algorithms}.
\newblock \emph{Neural Networks}, 12(1):145--151.

\bibitem[{Rae et~al.(2021)Rae, Borgeaud, Cai, Millican, Hoffmann, Song, Aslanides, Henderson, Ring, Young, Rutherford, Hennigan, Menick, Cassirer, Powell, van~den Driessche, Hendricks, Rauh, Huang, Glaese, Welbl, Dathathri, Huang, Uesato, Mellor, Higgins, Creswell, McAleese, Wu, Elsen, Jayakumar, Buchatskaya, Budden, Sutherland, Simonyan, Paganini, Sifre, Martens, Li, Kuncoro, Nematzadeh, Gribovskaya, Donato, Lazaridou, Mensch, Lespiau, Tsimpoukelli, Grigorev, Fritz, Sottiaux, Pajarskas, Pohlen, Gong, Toyama, de~Masson~d'Autume, Li, Terzi, Mikulik, Babuschkin, Clark, de~Las~Casas, Guy, Jones, Bradbury, Johnson, Hechtman, Weidinger, Gabriel, Isaac, Lockhart, Osindero, Rimell, Dyer, Vinyals, Ayoub, Stanway, Bennett, Hassabis, Kavukcuoglu, and Irving}]{gopher}
Jack~W. Rae, Sebastian Borgeaud, Trevor Cai, Katie Millican, Jordan Hoffmann, H.~Francis Song, John Aslanides, Sarah Henderson, Roman Ring, Susannah Young, Eliza Rutherford, Tom Hennigan, Jacob Menick, Albin Cassirer, Richard Powell, George van~den Driessche, Lisa~Anne Hendricks, Maribeth Rauh, Po{-}Sen Huang, Amelia Glaese, Johannes Welbl, Sumanth Dathathri, Saffron Huang, Jonathan Uesato, John Mellor, Irina Higgins, Antonia Creswell, Nat McAleese, Amy Wu, Erich Elsen, Siddhant~M. Jayakumar, Elena Buchatskaya, David Budden, Esme Sutherland, Karen Simonyan, Michela Paganini, Laurent Sifre, Lena Martens, Xiang~Lorraine Li, Adhiguna Kuncoro, Aida Nematzadeh, Elena Gribovskaya, Domenic Donato, Angeliki Lazaridou, Arthur Mensch, Jean{-}Baptiste Lespiau, Maria Tsimpoukelli, Nikolai Grigorev, Doug Fritz, Thibault Sottiaux, Mantas Pajarskas, Toby Pohlen, Zhitao Gong, Daniel Toyama, Cyprien de~Masson~d'Autume, Yujia Li, Tayfun Terzi, Vladimir Mikulik, Igor Babuschkin, Aidan Clark, Diego de~Las~Casas, Aurelia Guy,
  Chris Jones, James Bradbury, Matthew~J. Johnson, Blake~A. Hechtman, Laura Weidinger, Iason Gabriel, William Isaac, Edward Lockhart, Simon Osindero, Laura Rimell, Chris Dyer, Oriol Vinyals, Kareem Ayoub, Jeff Stanway, Lorrayne Bennett, Demis Hassabis, Koray Kavukcuoglu, and Geoffrey Irving. 2021.
\newblock \href {http://arxiv.org/abs/2112.11446} {Scaling language models: Methods, analysis {\&} insights from training gopher}.
\newblock \emph{CoRR}, abs/2112.11446.

\bibitem[{Raffel et~al.(2020)Raffel, Shazeer, Roberts, Lee, Narang, Matena, Zhou, Li, and Liu}]{t5c4}
Colin Raffel, Noam Shazeer, Adam Roberts, Katherine Lee, Sharan Narang, Michael Matena, Yanqi Zhou, Wei Li, and Peter~J. Liu. 2020.
\newblock \href {http://jmlr.org/papers/v21/20-074.html} {Exploring the limits of transfer learning with a unified text-to-text transformer}.
\newblock \emph{J. Mach. Learn. Res.}, 21:140:1--140:67.

\bibitem[{Rajbhandari et~al.(2020)Rajbhandari, Rasley, Ruwase, and He}]{zero}
Samyam Rajbhandari, Jeff Rasley, Olatunji Ruwase, and Yuxiong He. 2020.
\newblock \href {https://doi.org/10.1109/SC41405.2020.00024} {Zero: memory optimizations toward training trillion parameter models}.
\newblock In \emph{Proceedings of the International Conference for High Performance Computing, Networking, Storage and Analysis, {SC} 2020, Virtual Event / Atlanta, Georgia, USA, November 9-19, 2020}, page~20. {IEEE/ACM}.

\bibitem[{Ramesh et~al.(2021)Ramesh, Pavlov, Goh, Gray, Voss, Radford, Chen, and Sutskever}]{dalle}
Aditya Ramesh, Mikhail Pavlov, Gabriel Goh, Scott Gray, Chelsea Voss, Alec Radford, Mark Chen, and Ilya Sutskever. 2021.
\newblock \href {http://proceedings.mlr.press/v139/ramesh21a.html} {Zero-shot text-to-image generation}.
\newblock In \emph{Proceedings of the 38th International Conference on Machine Learning, {ICML} 2021, 18-24 July 2021, Virtual Event}, volume 139 of \emph{Proceedings of Machine Learning Research}, pages 8821--8831. {PMLR}.

\bibitem[{Ruder(2016)}]{optim_overview}
Sebastian Ruder. 2016.
\newblock \href {http://arxiv.org/abs/1609.04747} {An overview of gradient descent optimization algorithms}.
\newblock \emph{CoRR}, abs/1609.04747.

\bibitem[{Scao et~al.(2022)Scao, Fan, Akiki, Pavlick, Ilic, Hesslow, Castagn{\'{e}}, Luccioni, Yvon, Gall{\'{e}}, Tow, Rush, Biderman, Webson, Ammanamanchi, Wang, Sagot, Muennighoff, del Moral, Ruwase, Bawden, Bekman, McMillan{-}Major, Beltagy, Nguyen, Saulnier, Tan, Suarez, Sanh, Lauren{\c{c}}on, Jernite, Launay, Mitchell, Raffel, Gokaslan, Simhi, Soroa, Aji, Alfassy, Rogers, Nitzav, Xu, Mou, Emezue, Klamm, Leong, van Strien, Adelani, and et~al.}]{bloom}
Teven~Le Scao, Angela Fan, Christopher Akiki, Ellie Pavlick, Suzana Ilic, Daniel Hesslow, Roman Castagn{\'{e}}, Alexandra~Sasha Luccioni, Fran{\c{c}}ois Yvon, Matthias Gall{\'{e}}, Jonathan Tow, Alexander~M. Rush, Stella Biderman, Albert Webson, Pawan~Sasanka Ammanamanchi, Thomas Wang, Beno{\^{\i}}t Sagot, Niklas Muennighoff, Albert~Villanova del Moral, Olatunji Ruwase, Rachel Bawden, Stas Bekman, Angelina McMillan{-}Major, Iz~Beltagy, Huu Nguyen, Lucile Saulnier, Samson Tan, Pedro~Ortiz Suarez, Victor Sanh, Hugo Lauren{\c{c}}on, Yacine Jernite, Julien Launay, Margaret Mitchell, Colin Raffel, Aaron Gokaslan, Adi Simhi, Aitor Soroa, Alham~Fikri Aji, Amit Alfassy, Anna Rogers, Ariel~Kreisberg Nitzav, Canwen Xu, Chenghao Mou, Chris Emezue, Christopher Klamm, Colin Leong, Daniel van Strien, David~Ifeoluwa Adelani, and et~al. 2022.
\newblock \href {https://doi.org/10.48550/arXiv.2211.05100} {{BLOOM:} {A} 176b-parameter open-access multilingual language model}.
\newblock \emph{CoRR}, abs/2211.05100.

\bibitem[{Shazeer and Stern(2018)}]{adafactor}
Noam Shazeer and Mitchell Stern. 2018.
\newblock \href {http://proceedings.mlr.press/v80/shazeer18a.html} {Adafactor: Adaptive learning rates with sublinear memory cost}.
\newblock In \emph{Proceedings of the 35th International Conference on Machine Learning, {ICML} 2018, Stockholmsm{\"{a}}ssan, Stockholm, Sweden, July 10-15, 2018}, volume~80 of \emph{Proceedings of Machine Learning Research}, pages 4603--4611. {PMLR}.

\bibitem[{Sun et~al.(2022{\natexlab{a}})Sun, He, Qian, Zhou, Huang, and Qiu}]{bbtv2}
Tianxiang Sun, Zhengfu He, Hong Qian, Yunhua Zhou, Xuanjing Huang, and Xipeng Qiu. 2022{\natexlab{a}}.
\newblock \href {https://doi.org/10.18653/v1/2022.emnlp-main.259} {Bbtv2: Towards a gradient-free future with large language models}.
\newblock In \emph{Proceedings of the 2022 Conference on Empirical Methods in Natural Language Processing, {EMNLP} 2022, Abu Dhabi, United Arab Emirates, December 7-11, 2022}, pages 3916--3930. Association for Computational Linguistics.

\bibitem[{Sun et~al.(2022{\natexlab{b}})Sun, Shao, Qian, Huang, and Qiu}]{bbt}
Tianxiang Sun, Yunfan Shao, Hong Qian, Xuanjing Huang, and Xipeng Qiu. 2022{\natexlab{b}}.
\newblock \href {https://proceedings.mlr.press/v162/sun22e.html} {Black-box tuning for language-model-as-a-service}.
\newblock In \emph{International Conference on Machine Learning, {ICML} 2022, 17-23 July 2022, Baltimore, Maryland, {USA}}, volume 162 of \emph{Proceedings of Machine Learning Research}, pages 20841--20855. {PMLR}.

\bibitem[{Sun et~al.(2020)Sun, Wang, Chen, Ni, Agrawal, Cui, Venkataramani, Maghraoui, Srinivasan, and Gopalakrishnan}]{bit4optim}
Xiao Sun, Naigang Wang, Chia{-}Yu Chen, Jiamin Ni, Ankur Agrawal, Xiaodong Cui, Swagath Venkataramani, Kaoutar~El Maghraoui, Vijayalakshmi Srinivasan, and Kailash Gopalakrishnan. 2020.
\newblock \href {https://proceedings.neurips.cc/paper/2020/hash/13b919438259814cd5be8cb45877d577-Abstract.html} {Ultra-low precision 4-bit training of deep neural networks}.
\newblock In \emph{Advances in Neural Information Processing Systems 33: Annual Conference on Neural Information Processing Systems 2020, NeurIPS 2020, December 6-12, 2020, virtual}.

\bibitem[{Suzgun et~al.(2023)Suzgun, Scales, Sch{\"{a}}rli, Gehrmann, Tay, Chung, Chowdhery, Le, Chi, Zhou, and Wei}]{bbh}
Mirac Suzgun, Nathan Scales, Nathanael Sch{\"{a}}rli, Sebastian Gehrmann, Yi~Tay, Hyung~Won Chung, Aakanksha Chowdhery, Quoc~V. Le, Ed~Chi, Denny Zhou, and Jason Wei. 2023.
\newblock \href {https://aclanthology.org/2023.findings-acl.824} {Challenging big-bench tasks and whether chain-of-thought can solve them}.
\newblock In \emph{Findings of the Association for Computational Linguistics: {ACL} 2023, Toronto, Canada, July 9-14, 2023}, pages 13003--13051. Association for Computational Linguistics.

\bibitem[{Taori et~al.(2023)Taori, Gulrajani, Zhang, Dubois, Li, Guestrin, Liang, and Hashimoto}]{alpaca}
Rohan Taori, Ishaan Gulrajani, Tianyi Zhang, Yann Dubois, Xuechen Li, Carlos Guestrin, Percy Liang, and Tatsunori~B. Hashimoto. 2023.
\newblock Stanford alpaca: An instruction-following llama model.
\newblock \url{https://github.com/tatsu-lab/stanford_alpaca}.

\bibitem[{Touvron et~al.(2023{\natexlab{a}})Touvron, Lavril, Izacard, Martinet, Lachaux, Lacroix, Rozi{\`{e}}re, Goyal, Hambro, Azhar, Rodriguez, Joulin, Grave, and Lample}]{llama}
Hugo Touvron, Thibaut Lavril, Gautier Izacard, Xavier Martinet, Marie{-}Anne Lachaux, Timoth{\'{e}}e Lacroix, Baptiste Rozi{\`{e}}re, Naman Goyal, Eric Hambro, Faisal Azhar, Aur{\'{e}}lien Rodriguez, Armand Joulin, Edouard Grave, and Guillaume Lample. 2023{\natexlab{a}}.
\newblock \href {https://doi.org/10.48550/arXiv.2302.13971} {Llama: Open and efficient foundation language models}.
\newblock \emph{CoRR}, abs/2302.13971.

\bibitem[{Touvron et~al.(2023{\natexlab{b}})Touvron, Martin, Stone, Albert, Almahairi, Babaei, Bashlykov, Batra, Bhargava, Bhosale, Bikel, Blecher, Canton{-}Ferrer, Chen, Cucurull, Esiobu, Fernandes, Fu, Fu, Fuller, Gao, Goswami, Goyal, Hartshorn, Hosseini, Hou, Inan, Kardas, Kerkez, Khabsa, Kloumann, Korenev, Koura, Lachaux, Lavril, Lee, Liskovich, Lu, Mao, Martinet, Mihaylov, Mishra, Molybog, Nie, Poulton, Reizenstein, Rungta, Saladi, Schelten, Silva, Smith, Subramanian, Tan, Tang, Taylor, Williams, Kuan, Xu, Yan, Zarov, Zhang, Fan, Kambadur, Narang, Rodriguez, Stojnic, Edunov, and Scialom}]{llama2}
Hugo Touvron, Louis Martin, Kevin Stone, Peter Albert, Amjad Almahairi, Yasmine Babaei, Nikolay Bashlykov, Soumya Batra, Prajjwal Bhargava, Shruti Bhosale, Dan Bikel, Lukas Blecher, Cristian Canton{-}Ferrer, Moya Chen, Guillem Cucurull, David Esiobu, Jude Fernandes, Jeremy Fu, Wenyin Fu, Brian Fuller, Cynthia Gao, Vedanuj Goswami, Naman Goyal, Anthony Hartshorn, Saghar Hosseini, Rui Hou, Hakan Inan, Marcin Kardas, Viktor Kerkez, Madian Khabsa, Isabel Kloumann, Artem Korenev, Punit~Singh Koura, Marie{-}Anne Lachaux, Thibaut Lavril, Jenya Lee, Diana Liskovich, Yinghai Lu, Yuning Mao, Xavier Martinet, Todor Mihaylov, Pushkar Mishra, Igor Molybog, Yixin Nie, Andrew Poulton, Jeremy Reizenstein, Rashi Rungta, Kalyan Saladi, Alan Schelten, Ruan Silva, Eric~Michael Smith, Ranjan Subramanian, Xiaoqing~Ellen Tan, Binh Tang, Ross Taylor, Adina Williams, Jian~Xiang Kuan, Puxin Xu, Zheng Yan, Iliyan Zarov, Yuchen Zhang, Angela Fan, Melanie Kambadur, Sharan Narang, Aur{\'{e}}lien Rodriguez, Robert Stojnic, Sergey Edunov,
  and Thomas Scialom. 2023{\natexlab{b}}.
\newblock \href {https://doi.org/10.48550/arXiv.2307.09288} {Llama 2: Open foundation and fine-tuned chat models}.
\newblock \emph{CoRR}, abs/2307.09288.

\bibitem[{Vaswani et~al.(2017)Vaswani, Shazeer, Parmar, Uszkoreit, Jones, Gomez, Kaiser, and Polosukhin}]{transformer}
Ashish Vaswani, Noam Shazeer, Niki Parmar, Jakob Uszkoreit, Llion Jones, Aidan~N Gomez, {\L}ukasz Kaiser, and Illia Polosukhin. 2017.
\newblock Attention is all you need.
\newblock \emph{Advances in neural information processing systems}, 30.

\bibitem[{Wang et~al.(2019)Wang, Pruksachatkun, Nangia, Singh, Michael, Hill, Levy, and Bowman}]{superglue}
Alex Wang, Yada Pruksachatkun, Nikita Nangia, Amanpreet Singh, Julian Michael, Felix Hill, Omer Levy, and Samuel~R. Bowman. 2019.
\newblock \href {http://arxiv.org/abs/1905.00537} {Superglue: {A} stickier benchmark for general-purpose language understanding systems}.
\newblock \emph{CoRR}, abs/1905.00537.

\bibitem[{Wang et~al.(2023)Wang, Kordi, Mishra, Liu, Smith, Khashabi, and Hajishirzi}]{self-instruct}
Yizhong Wang, Yeganeh Kordi, Swaroop Mishra, Alisa Liu, Noah~A. Smith, Daniel Khashabi, and Hannaneh Hajishirzi. 2023.
\newblock \href {https://doi.org/10.18653/v1/2023.acl-long.754} {Self-instruct: Aligning language models with self-generated instructions}.
\newblock In \emph{Proceedings of the 61st Annual Meeting of the Association for Computational Linguistics (Volume 1: Long Papers), {ACL} 2023, Toronto, Canada, July 9-14, 2023}, pages 13484--13508. Association for Computational Linguistics.

\bibitem[{Yu et~al.(2018)Yu, Zhou, Cichocki, and Xie}]{matrix_factor}
Jinshi Yu, Guoxu Zhou, Andrzej Cichocki, and Shengli Xie. 2018.
\newblock \href {https://doi.org/10.1109/ACCESS.2018.2873385} {Learning the hierarchical parts of objects by deep non-smooth nonnegative matrix factorization}.
\newblock \emph{{IEEE} Access}, 6:58096--58105.

\bibitem[{Zeiler(2012)}]{adadelta}
Matthew~D. Zeiler. 2012.
\newblock \href {http://arxiv.org/abs/1212.5701} {{ADADELTA:} an adaptive learning rate method}.
\newblock \emph{CoRR}, abs/1212.5701.

\bibitem[{Zhang et~al.(2024)Zhang, Zeng, Wang, and Lu}]{zhang2024tinyllama}
Peiyuan Zhang, Guangtao Zeng, Tianduo Wang, and Wei Lu. 2024.
\newblock \href {http://arxiv.org/abs/2401.02385} {Tinyllama: An open-source small language model}.

\bibitem[{Zhang et~al.(2022)Zhang, Roller, Goyal, Artetxe, Chen, Chen, Dewan, Diab, Li, Lin, Mihaylov, Ott, Shleifer, Shuster, Simig, Koura, Sridhar, Wang, and Zettlemoyer}]{opt}
Susan Zhang, Stephen Roller, Naman Goyal, Mikel Artetxe, Moya Chen, Shuohui Chen, Christopher Dewan, Mona~T. Diab, Xian Li, Xi~Victoria Lin, Todor Mihaylov, Myle Ott, Sam Shleifer, Kurt Shuster, Daniel Simig, Punit~Singh Koura, Anjali Sridhar, Tianlu Wang, and Luke Zettlemoyer. 2022.
\newblock \href {https://doi.org/10.48550/arXiv.2205.01068} {{OPT:} open pre-trained transformer language models}.
\newblock \emph{CoRR}, abs/2205.01068.

\end{thebibliography}
\newpage
\appendix

\section{Empirical Analysis on the Two Moments}
\label{sec:appendix_analysis_moments}
We also empirically investigated the differences in convergence behaviors between Adam and SGD under the function \(f(x,y) = x^2 + y^2 - 2 e^{-5[(x-1)^2+y^2]} - 3 e^{-5[(x+1)^2 + y^2]}\).

\begin{figure}[h]
    \centering
    \includegraphics[width=1\linewidth]{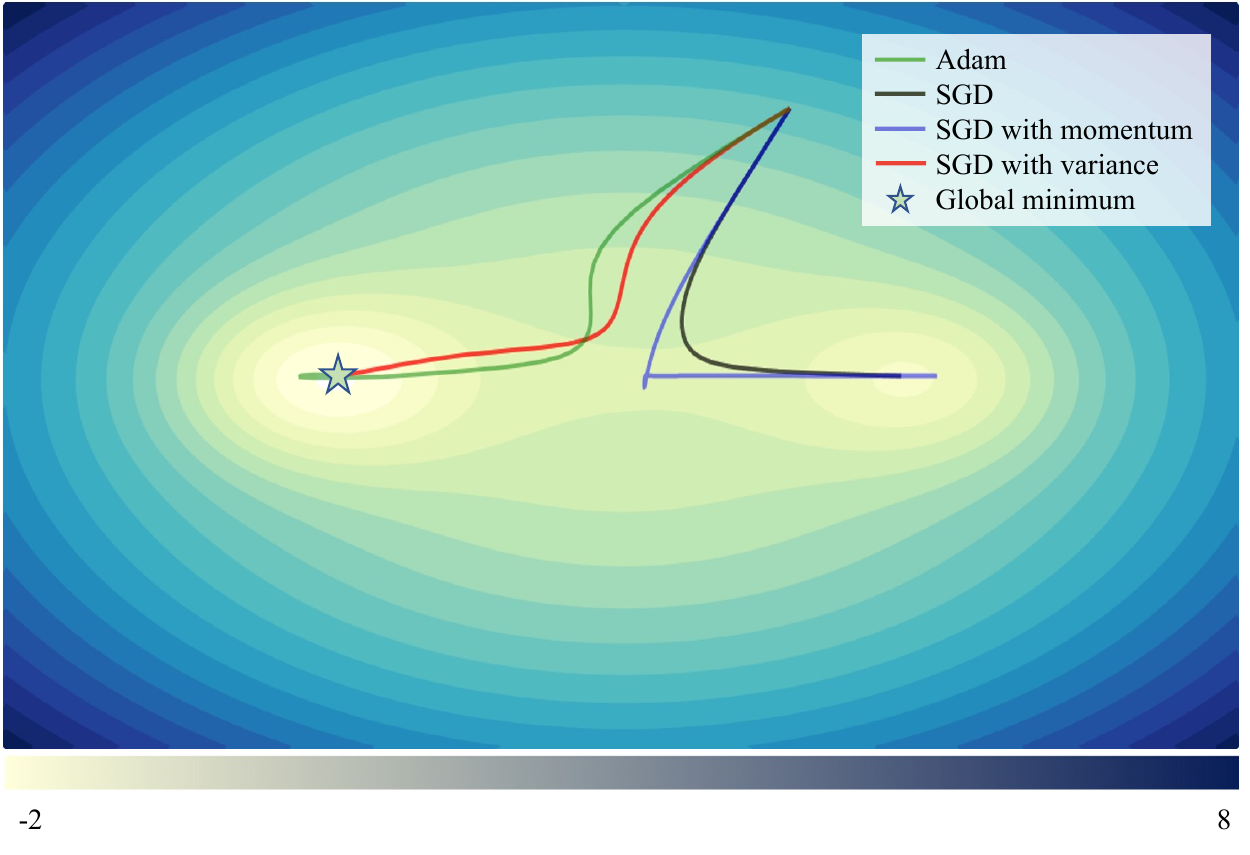}
        \caption{Empirical analysis on different optimization methods. Loss trajectories of different optimizers starting from the same initial point. Both Adam and SGD with variance converge to the global optimum on the left, while SGD and SGD with momentum converge to the local optimum on the right.}
        \label{fig:optim_function}
\end{figure}

The results of the convergence analysis are shown in Figure~\ref{fig:optim_function}. Starting from the same initial point, Adam converges to the global optimum while SGD gets trapped at a local optimum. 

\section{Gradient Normalization for AdaLomo}
\label{appendix:grad_norm}

\begin{figure*}
    \centering
    \begin{subfigure}{0.32\textwidth}
        \centering
        \includegraphics[width=\textwidth]{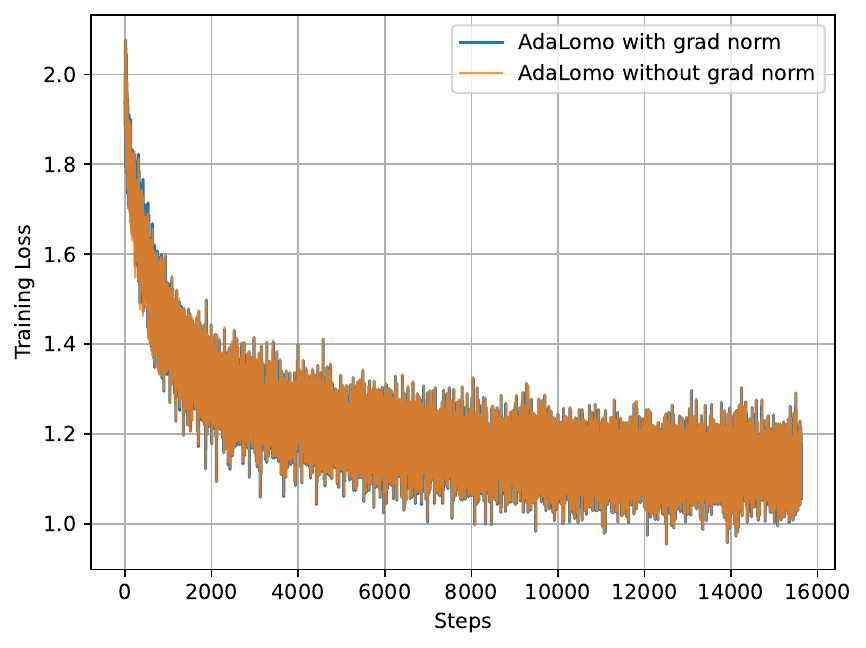}
        \caption{Training loss curve.}
    \end{subfigure}
    \hfill
    \begin{subfigure}{0.32\textwidth}
        \centering
        \includegraphics[width=\textwidth]{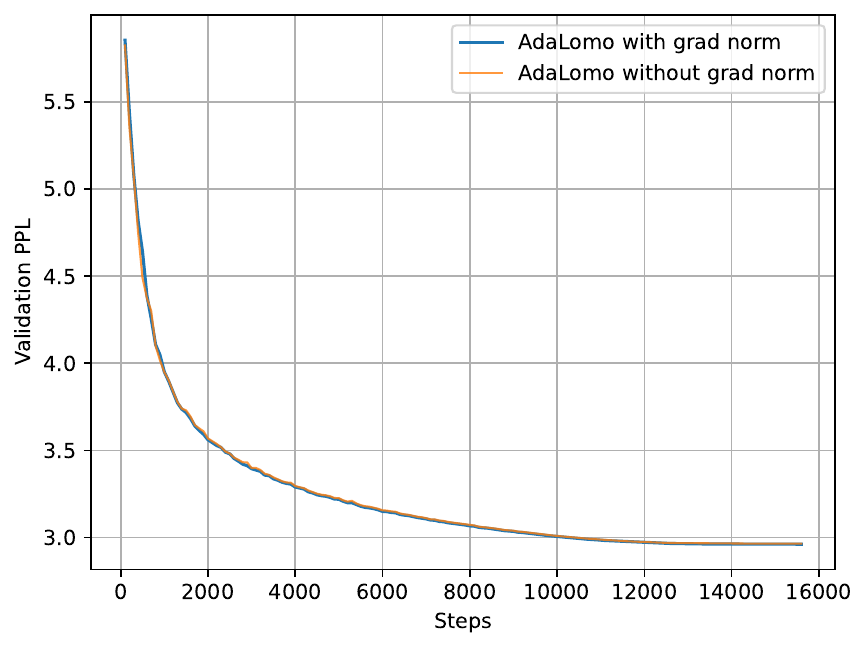}
        \caption{Validation perplexity.}
    \end{subfigure}
    \hfill
    \begin{subfigure}{0.32\textwidth}
        \centering
        \includegraphics[width=\textwidth]{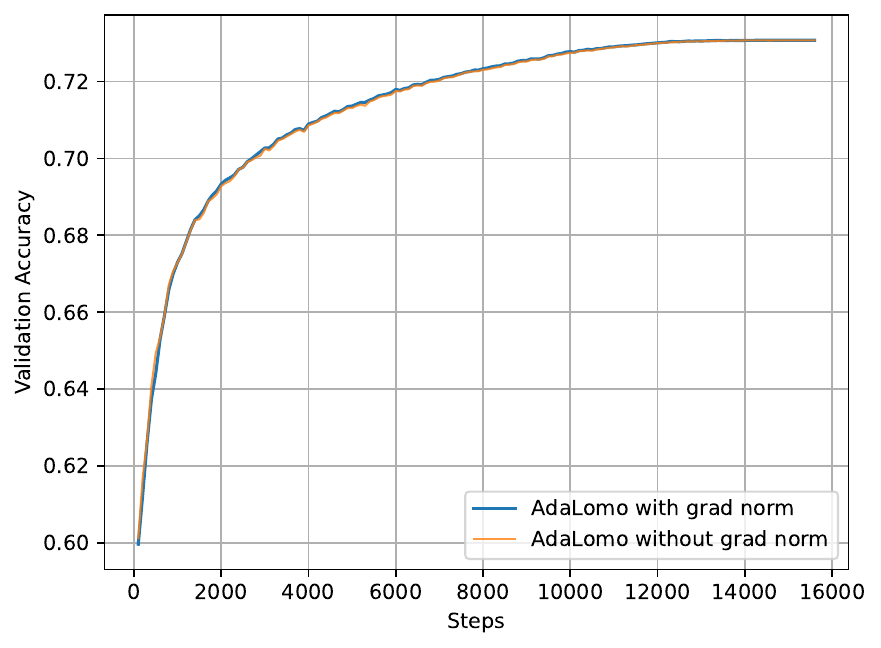}
        \caption{Validation next-token accuracy.}
    \end{subfigure}
    \caption{Results of further pre-training of LLaMA-7B with AdaLomo in the Chinese domain with and without gradient normalization.}
    \label{fig:cn_grad_norm}
\end{figure*}

\begin{figure*}
    \centering
    \begin{subfigure}{0.32\textwidth}
        \centering
        \includegraphics[width=\textwidth]{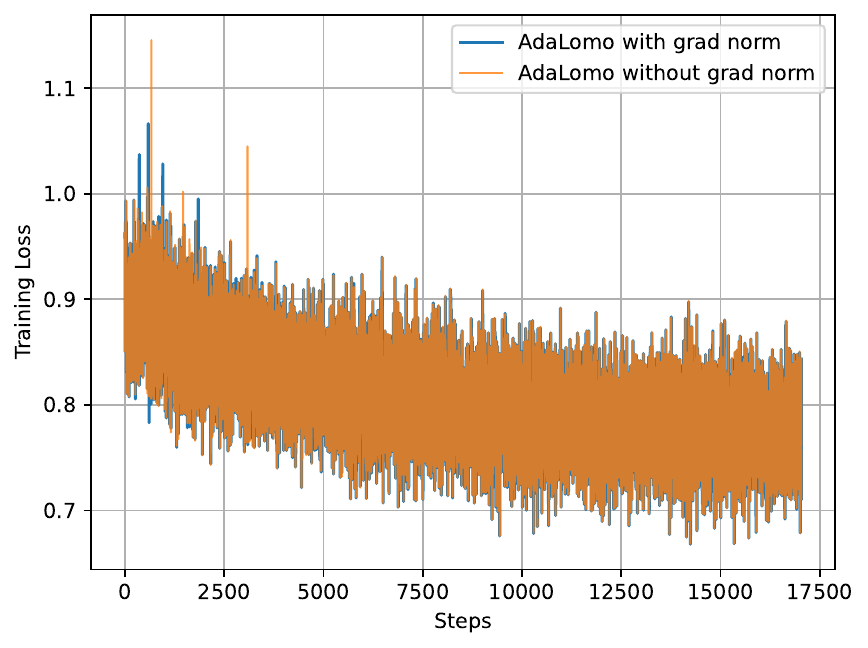}
        \caption{Training loss curve.}
    \end{subfigure}
    \hfill
    \begin{subfigure}{0.32\textwidth}
        \centering
        \includegraphics[width=\textwidth]{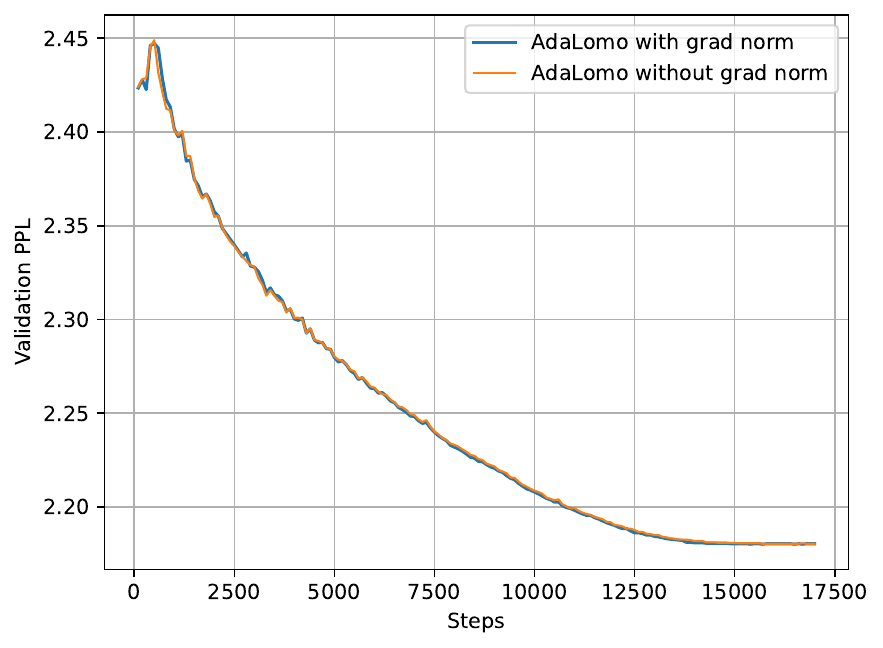}
        \caption{Validation perplexity.}
    \end{subfigure}
    \hfill
    \begin{subfigure}{0.32\textwidth}
        \centering
        \includegraphics[width=\textwidth]{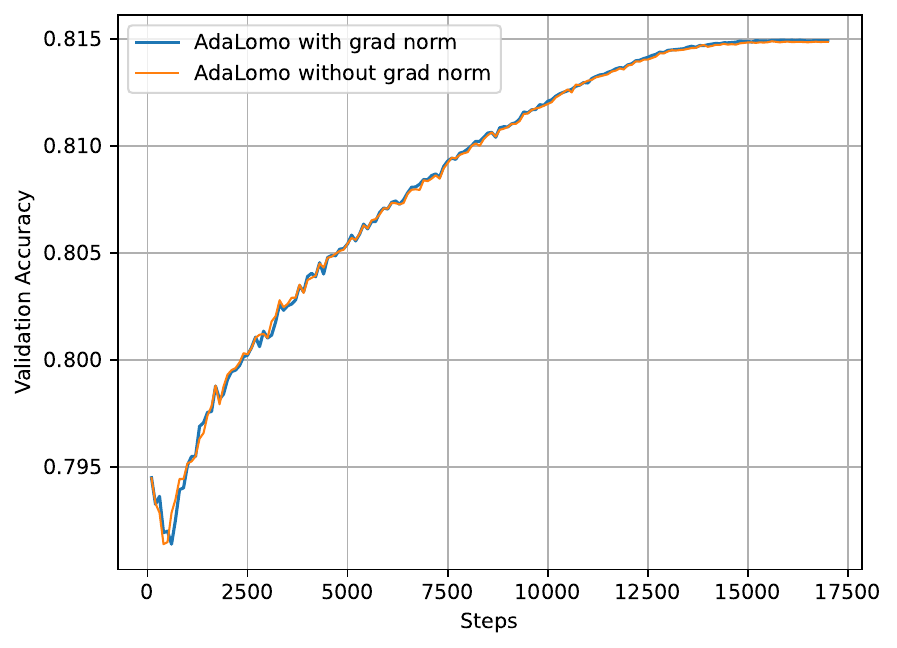}
        \caption{Validation next-token accuracy.}
    \end{subfigure}
    \caption{Results of further pre-training of LLaMA-7B with AdaLomo in the Python code domain with and without gradient normalization.}
    \label{fig:py_grad_norm}
\end{figure*}

We conduct experiments on the LLaMA-7B to assess the effects of using gradient normalization during the further pre-training of AdaLomo. Comparative experiments in the Chinese domain are illustrated in Figure~\ref{fig:cn_grad_norm}, while those in the Python code domain are shown in Figure~\ref{fig:py_grad_norm}. Our results indicate that the convergence performance of AdaLomo is unaffected by the use or absence of gradient normalization. We attribute this to the grouped update normalization feature within AdaLomo. Avoiding the use of gradient normalization can eliminate the need for two backward passes, thus preventing computational redundancy during training.

\section{Instruction Tuning}
\label{appendix:instruction-tuning}
\subsection{Hyper-parameters}
Hyper-parameters used by different optimization methods and models for instruction-tuning are shown in Table~\ref{tab:hyper-param-sft}.

\begin{table*}[ht]
\small
\centering
\begin{tabular}{ccccccccc}
\toprule
\multirow{2}{*}{\textbf{}}                              & \multicolumn{4}{c}{LLaMA-7B}    & \multicolumn{4}{c}{LLaMA-13B}   \\ \cline{2-9} 
                                                        & LoRA  & AdamW & LOMO  & AdaLomo & LoRA  & AdamW & LOMO  & AdaLomo \\ \hline
\begin{tabular}[c]{@{}c@{}}Learning\\ Rate\end{tabular} & 3E-04 & 2E-05 & 1E-02 & 5E-04   & 3E-04 & 2E-05 & 1E-02 & 5E-04   \\ \midrule
\begin{tabular}[c]{@{}c@{}}Batch\\ Size\end{tabular}    & \multicolumn{8}{c}{128}                                           \\ \midrule
Ecochs                                                  & \multicolumn{8}{c}{3}                                             \\ \midrule
\begin{tabular}[c]{@{}c@{}}Warmup\\ Steps\end{tabular}  & \multicolumn{8}{c}{0.03 * Total Steps}                            \\ \midrule
                                                        & \multicolumn{4}{c}{LLaMA-30B}   & \multicolumn{4}{c}{LLaMA-65B}   \\ \cline{2-9} 
                                                        & LoRA  & AdamW & LOMO  & AdaLomo & LoRA  & AdamW & LOMO  & AdaLomo \\ \hline
\begin{tabular}[c]{@{}c@{}}Learning\\ Rate\end{tabular} & 3E-04 & 2E-05 & 1E-02 & 5E-04   & 3E-04 & 1E-05 & 1E-02 & 5E-04   \\ \midrule
\begin{tabular}[c]{@{}c@{}}Batch\\ Size\end{tabular}    & \multicolumn{8}{c}{128}                                           \\ \midrule
Ecochs                                                  & \multicolumn{8}{c}{3}                                             \\ \midrule
\begin{tabular}[c]{@{}c@{}}Warmup\\ Steps\end{tabular}  & \multicolumn{8}{c}{0.03 * Total Steps}                            \\ \bottomrule
\end{tabular}
\caption{Hyper-parameters for instruction-tuning.}
\label{tab:hyper-param-sft}
\end{table*}

\subsection{Templates}
Templates used for instruction-tuning on Alpaca-GPT4 are shown in Table~\ref{tab:template}.

\begin{table*}
\centering
\begin{tabularx}{\textwidth}{X}
  \toprule
\textbf{Template for entries with input} \\
\midrule
Below is an instruction that describes a task, paired with an input that provides further context. Write a response that appropriately completes the request.\\\\
\#\#\# Instruction:\\
\{instruction\}\\\\
\#\#\# Input:\\
\{input\}\\\\
\#\#\# Response:\{response\}\\
\toprule
\textbf{Template for entries without input} \\
\midrule
Below is an instruction that describes a task. Write a response that appropriately completes the request.\\\\
\#\#\# Instruction:\\
\{instruction\}\\\\
\#\#\# Response:\{response\} \\
\bottomrule
\end{tabularx}
\caption{Templates used for instruction-tuning.}
\label{tab:template}
\label{tab:tem_train}
\end{table*}

\subsection{More Results}
In Table~\ref{table:more_results_sft}, we include a comparison of Adafactor on LLaMA-7B. The results show that Adafactor's performance is similar to AdaLomo's. Both Adafactor and AdaLomo significantly outperform LOMO on instruction-following task (AlpacaFarm).

\begin{table*}[]
\centering
\begin{tabular}{@{}lllllll@{}}
\toprule
Model     & MMLU & BBH  & GSM8K & HumanEval & AlpacaFarm & Avg. \\ \midrule
LLaMA-7B  & 31.5 & 32.3 & 10.9  & 11.6      & 4.2        & 18.1 \\
LoRA      & 33.5 & 34.8 & 12.3  & 11.0      & 41.1       & 26.5 \\
AdamW     & 39.3 & 34.4 & 9.66  & 11.6     & 50.6       & 29.1 \\
LOMO      & 30.7 & 34.0 & 12.0  & \textbf{12.8}      & 30.6       & 24.0 \\
Adafactor & \textbf{40.8} & 35.8 & \textbf{14.9}  & 11.0      & 47.7       & 30.0 \\
\rowcolor[gray]{0.9} AdaLomo   & 39.5 & \textbf{36.0} & 14.4  & 11.0      & \textbf{53.3}       & \textbf{30.8} \\ \bottomrule
\end{tabular}
\caption{Performance of the LLaMA-7B after instruction-tuning with different
optimization techniques.}
\label{table:more_results_sft}
\end{table*}

\section{Further Pre-training}
\label{appendix:further-train}

\subsection{Hyper-parameters}
Hyper-parameters used for further pre-trianing are shown in Table~\ref{tab:hyper-param-pretrain}.
\begin{table}[ht]
\centering
\begin{tabular}{@{}ccc@{}}
\toprule
Method          & AdamW             & AdaLomo            \\ \midrule
Sequence Length & \multicolumn{2}{c}{2048}               \\
Learning Rate   & 1E-05             & 3E-01              \\
Batch Size      & \multicolumn{2}{c}{128}                \\
Warmup Steps    & \multicolumn{2}{c}{0.03 * Total Steps} \\ \bottomrule
\end{tabular}
\caption{Hyper-parameters used for further pre-training.}
\label{tab:hyper-param-pretrain}
\end{table}

\subsection{More Results}
\begin{figure*}
    \centering
    \begin{subfigure}{0.32\textwidth}
        \centering
        \includegraphics[height=0.71\textwidth]{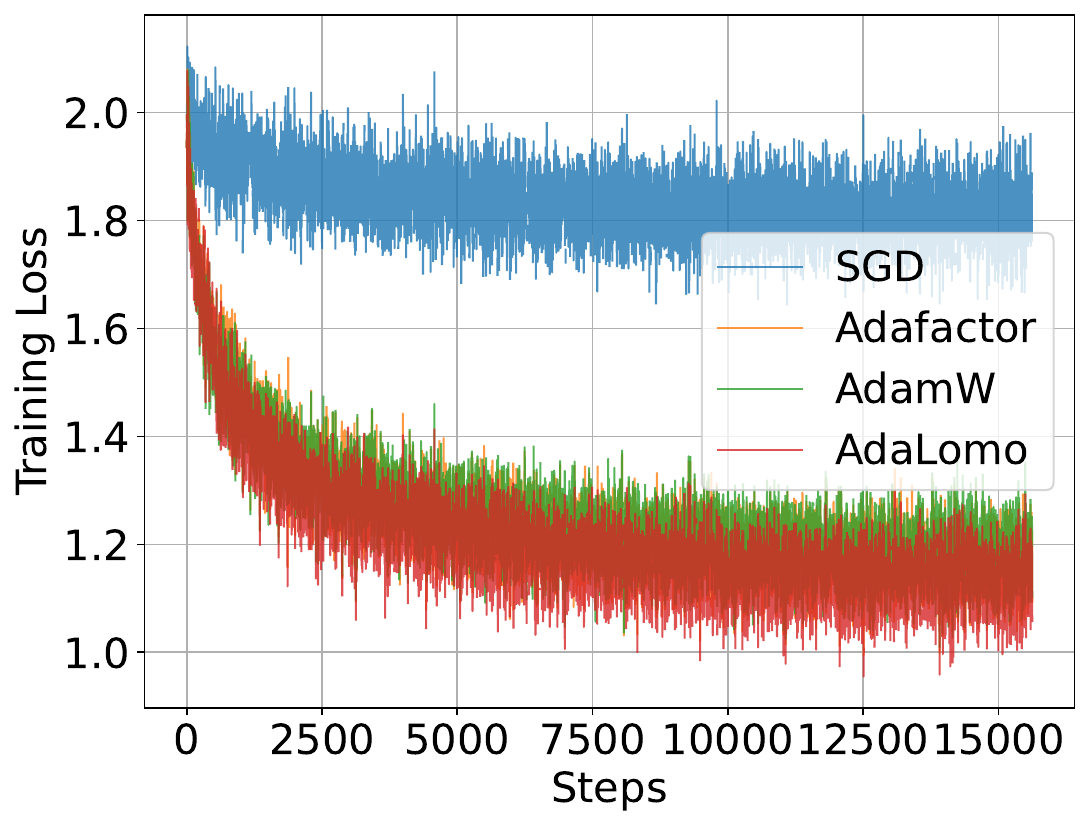}
        \caption{Training loss curve.}
    \end{subfigure}
    \hfill
    \begin{subfigure}{0.32\textwidth}
        \centering
        \includegraphics[height=0.71\textwidth]{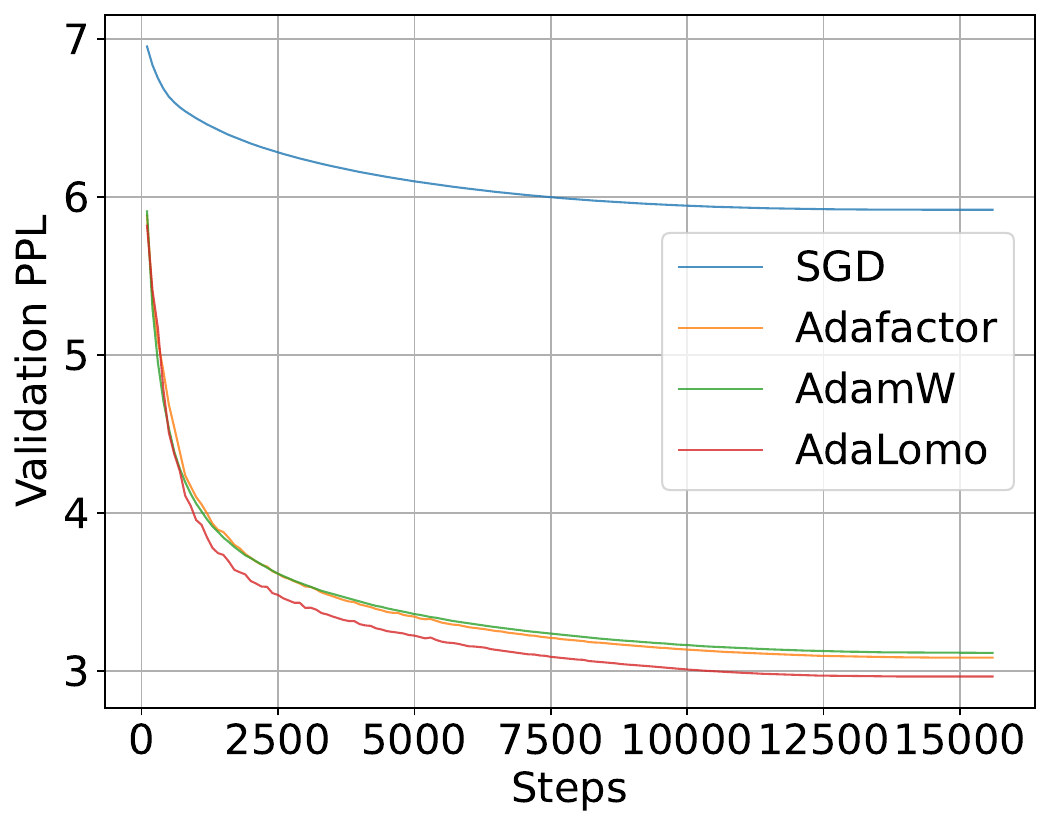}
        \caption{Validation perplexity.}
    \end{subfigure}
    \hfill
    \begin{subfigure}{0.32\textwidth}
        \centering
        \includegraphics[height=0.71\textwidth]{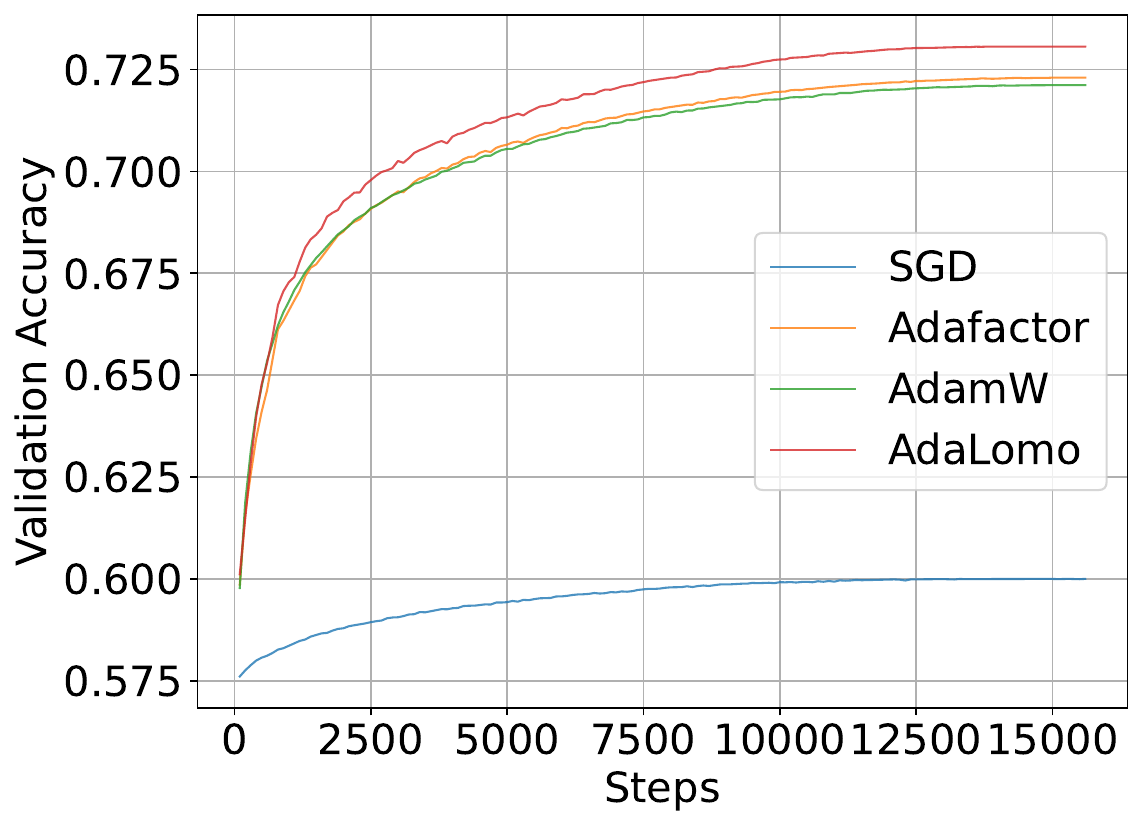}
        \caption{Validation next-token accuracy.}
    \end{subfigure}
    \caption{Results of further pre-training in the Chinese domain.}
    \label{fig:more_further_cn}
\end{figure*}

\begin{figure*}
    \centering
    \begin{subfigure}{0.32\textwidth}
        \centering
        \includegraphics[height=0.71\textwidth]{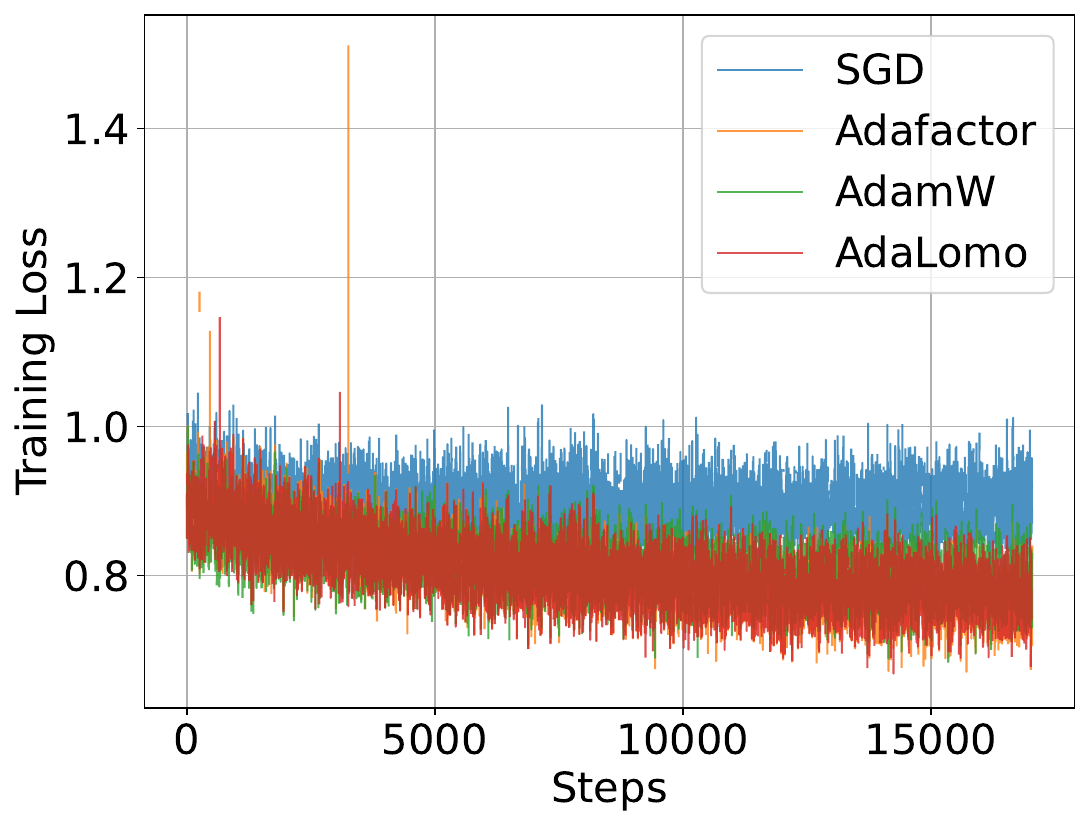}
        \caption{Training loss curve.}
    \end{subfigure}
    \hfill
    \begin{subfigure}{0.32\textwidth}
        \centering
        \includegraphics[height=0.71\textwidth]{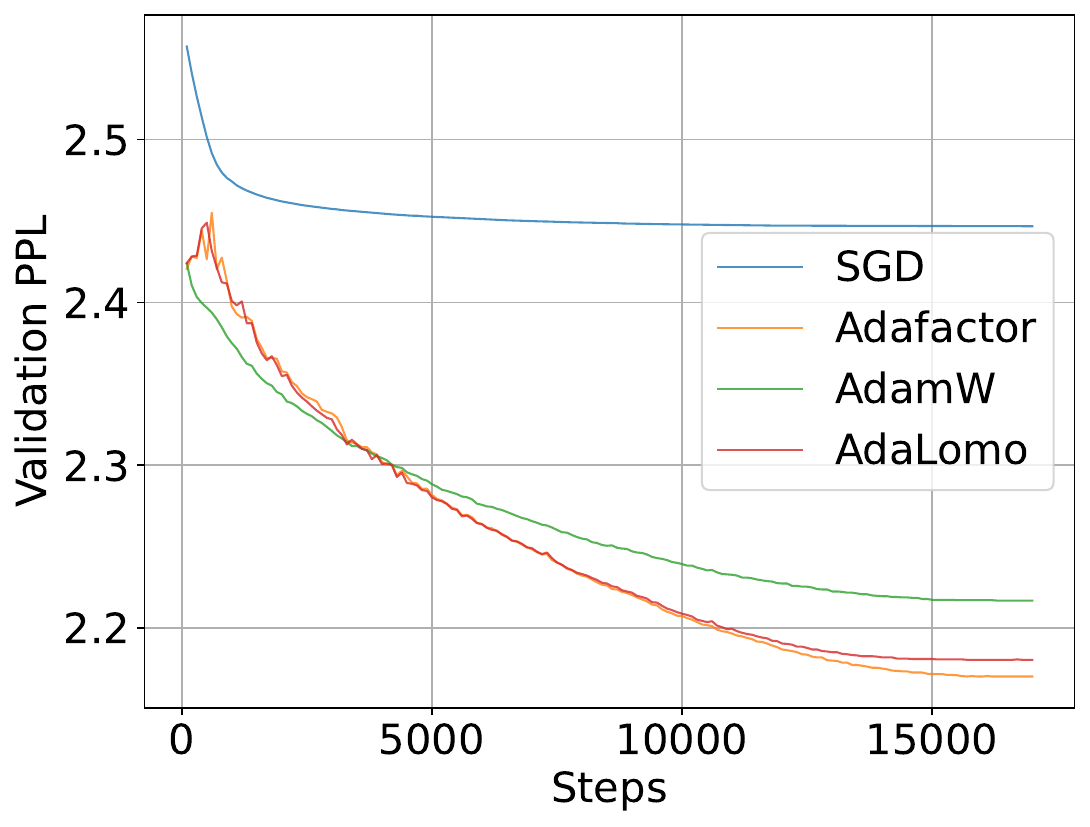}
        \caption{Validation perplexity.}
    \end{subfigure}
    \hfill
    \begin{subfigure}{0.32\textwidth}
        \centering
        \includegraphics[height=0.71\textwidth]{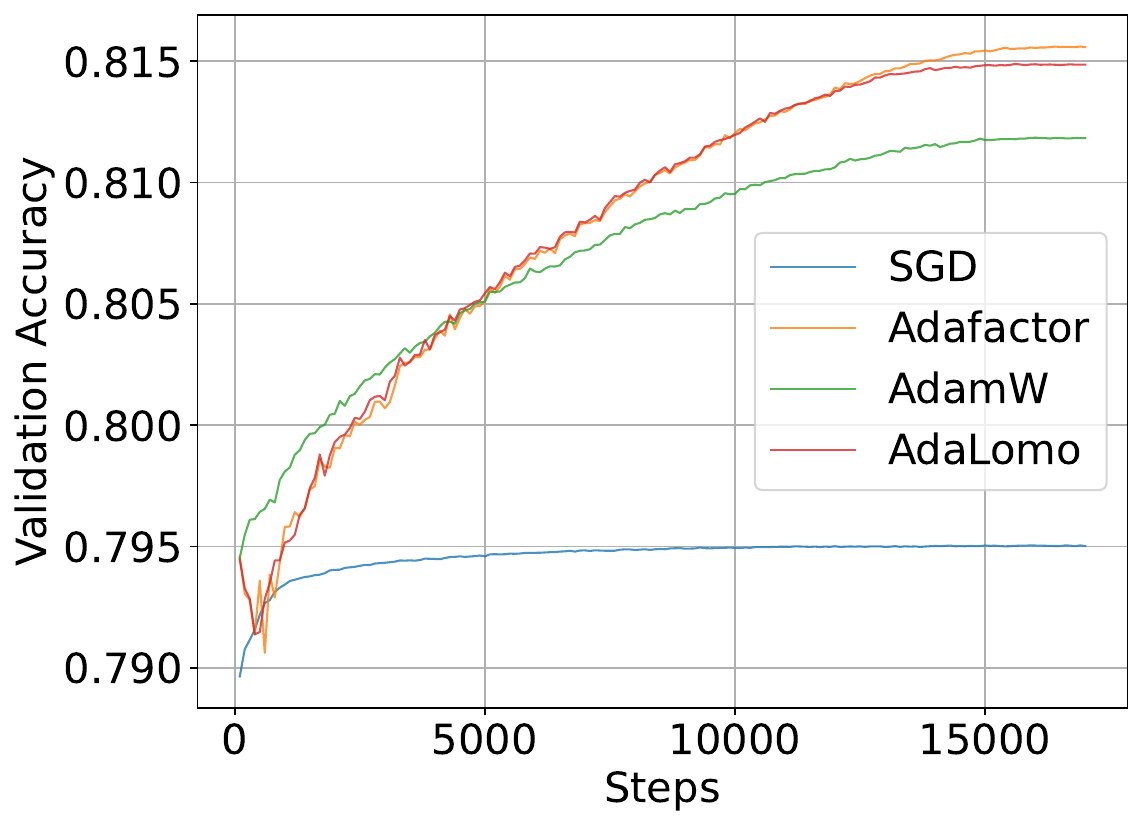}
        \caption{Validation next-token accuracy.}
    \end{subfigure}
    \caption{Results of further pre-training in the Python code domain.}
    \label{fig:more_further_py}
\end{figure*}

We present the results of further pre-training in the Chinese domain and the Python code domain on the LLaMA-7B model in Figure~\ref{fig:more_further_cn} and Figure~\ref{fig:more_further_py}, respectively. It can be observed that AdaLomo, AdamW, and Adafactor exhibit similar convergence speeds and final performance, while SGD performs poorly in both domains. This experiment confirms our hypothesis: second-order moments are crucial for optimizing transformer-based large language models.

\section{Pre-training from Scratch}
Our experimental comparisons and learning rates are shown in Table~\ref{tab:hyper-parameter-pretrain}, with AdamW's weight decay set to 0.01. 

\begin{table}
    \centering
    \small
    \begin{tabular}{ccccc}
    \toprule
         & SGD & Adafactor & AdamW & AdaLomo \\
         \midrule
        LR & 1e-3 & 1e-3 & 2e-5 & 1e-3 \\
    \bottomrule
    \end{tabular}
    \caption{Hyper-parameters for pre-training from scratch.}
    \label{tab:hyper-parameter-pretrain}
\end{table}

\section{Memory and Throughput Profile}
\label{appendix:profile}

The hyper-parameters used to profile memory and throughput and the detailed results are shown in Table~\ref{tab:profile}. The experiments are conducted on A800 with NVLink. For practical scenarios, we employ pynvml (Python NVIDIA Management Library) to record system-level memory usage.

\begin{table*}[ht]
\centering
\begin{tabular}{@{}cccccc@{}}
\toprule
Model                      & Optimizer & GPUs                & Micro Batch Size   & Memory (GB) & Throughput (TGS) \\ \midrule
\multirow{5}{*}{LLaMA-7B}  & AdamW     & \multirow{5}{*}{4}  & \multirow{5}{*}{8} & 169.4  & 3169.4     \\
                           & Adafactor &                     &                    & 144.3  & 3169.5     \\
                           & LoRA      &                     &                    & 70.6   & 3344.6     \\
                           & LOMO      &                     &                    & 59.6   & 3228.2     \\
                           & AdaLomo   &                     &                    & 59.6   & 2997.4     \\ \midrule
\multirow{5}{*}{LLaMA-30B} & AdamW     & \multirow{5}{*}{16} & \multirow{5}{*}{4} & 786.2  & 728.6      \\
                           & Adafactor &                     &                    & 665.0  & 726.5      \\
                           & LoRA      &                     &                    & 303.7  & 811.6      \\
                           & LOMO      &                     &                    & 264.3  & 669.1      \\
                           & AdaLomo   &                     &                    & 272.8  & 589.0      \\ \midrule
\multirow{5}{*}{LLaMA-13B} & AdamW     & \multirow{5}{*}{8}  & \multirow{5}{*}{4} & 320.7  & 1679.6     \\
                           & Adafactor &                     &                    & 272.3  & 1683.4     \\
                           & LoRA      &                     &                    & 110.0  & 1829.8     \\
                           & LOMO      &                     &                    & 94.4   & 1659.9     \\
                           & AdaLomo   &                     &                    & 95.8   & 1456.3     \\ \midrule
\multirow{5}{*}{LLaMA-65B} & AdamW     & \multirow{5}{*}{32} & \multirow{5}{*}{2} & 1532.6 & 349.1      \\
                           & Adafactor &                     &                    & 1289.4 & 341.1      \\
                           & LoRA      &                     &                    & 510.5  & 405.7      \\
                           & LOMO      &                     &                    & 473.8  & 303.3      \\
                           & AdaLomo   &                     &                    & 507.7  & 238.1      \\ \bottomrule
\end{tabular}
\caption{Hyper-parameters and detailed results in memory and throughput profile.}
\label{tab:profile}
\end{table*}

\end{document}